\pdfoutput=1

\documentclass[11pt]{article}

\usepackage[final]{acl}

\usepackage{times}
\usepackage{latexsym}
\usepackage{hyperref}
\usepackage[T1]{fontenc}

\usepackage[utf8]{inputenc}

\usepackage{microtype}

\usepackage{inconsolata}

\usepackage{graphicx}

\usepackage{amsmath}  
\usepackage{float}
\usepackage{tabularx}
\usepackage{array}
\usepackage{subcaption}
\usepackage{xcolor}
\usepackage{multirow}
\usepackage{booktabs} 
\usepackage{longtable}
\usepackage{colortbl}
\usepackage{graphicx}
\usepackage{etoolbox}

\usepackage{amssymb}
\definecolor{PineGreen}{RGB}{1, 121, 111}

\title{Multimodal Language Models See Better When They Look Shallower}

\author{
 \textbf{Haoran Chen\textsuperscript{1,2,3}}\thanks{Equal contribution.},
 \textbf{Junyan Lin\textsuperscript{2,3}}\footnotemark[1],
 \textbf{Xinghao Chen\textsuperscript{2,3}},
 \textbf{Yue Fan\textsuperscript{4}},
\\
 \textbf{Jianfeng Dong\textsuperscript{1}},
 \textbf{Xin Jin\textsuperscript{2,3}},
 \textbf{Hui Su\textsuperscript{5}},
 \textbf{Jinlan Fu\textsuperscript{6}},
 \textbf{Xiaoyu Shen\textsuperscript{2,3}}\thanks{Corresponding Author. Code available at \url{https://github.com/EIT-NLP/VisualProbing-for-MLLM}.}
\\
 \textsuperscript{1}Zhejiang Gongshang University
\\
  \textsuperscript{2}Ningbo Key Laboratory of Spatial Intelligence and Digital Derivative
  \\
  \textsuperscript{3}Institute of Digital Twin, Eastern Institute of Technology, Ningbo
\\
 \textsuperscript{4}Genmo.ai,
 \textsuperscript{5}Meituan Inc.,
 \textsuperscript{6}National University of Singapore
\\
 \small{
  \href{mailto:email@domain}{haoranchr321@gmail.com}, \href{mailto:email@domain}{xyshen@eitech.edu.cn}
 }
}

\begin{document}

\maketitle
\begin{abstract}

Multimodal large language models (MLLMs) typically extract visual features from the final layers of a pretrained Vision Transformer (ViT). This widespread deep-layer bias, however, is largely driven by empirical convention rather than principled analysis. While prior studies suggest that different ViT layers capture different types of information—shallower layers focusing on fine visual details and deeper layers aligning more closely with textual semantics, the impact of this variation on MLLM performance remains underexplored. We present the first comprehensive study of visual layer selection for MLLMs, analyzing representation similarity across ViT layers to establish \underline{shallow, middle, and deep} layer groupings. Through extensive evaluation of MLLMs (1.4B–7B parameters) across 10  benchmarks encompassing 60+ tasks, we find that while deep layers excel in semantic-rich tasks like OCR, shallow and middle layers significantly outperform them on fine-grained visual tasks including counting, positioning, and object localization. Building on these insights, we propose a lightweight feature fusion method that strategically incorporates shallower layers, achieving consistent improvements over both single-layer and specialized fusion baselines. Our work offers the first principled study of visual layer selection in MLLMs, showing that \emph{MLLMs can often see better when they look shallower}.

\end{abstract}

\section{Introduction}
\label{sec:intro}
Multimodal Large Language Models (MLLMs) extend the capabilities of traditional Large Language Models (LLMs) by enabling joint reasoning over both visual and textual inputs~\cite{hong2024cogagent, bai2023qwen, chen2024evlmefficientvisionlanguagemodel,shen2024zoomeye}. Typically, these models integrate a pretrained Vision Transformer (ViT) to extract image features, which are then projected into the language embedding space of an LLM. This architecture enables unified multimodal understanding and powers a wide range of applications, including robotic navigation, medical diagnostics, and visual question answering~\cite{alayrac2022flamingo, chen2024internvl, bai2023qwen, tong2024cambrian,li2025m2iv}.

While recent advancements have notably improved the language reasoning capabilities of MLLMs, the visual processing pipeline, particularly the selection of ViT layers for visual representation, remains insufficiently explored. In practice, MLLMs often default to using features from the deepest layers of ViT models. For instance, \texttt{Qwen-VL}~\cite{bai2025qwen2} and \texttt{InternVL-6B v1.2/1.5} use the final layer of CLIP-ViT~\cite{radford2021learning}, while other \texttt{InternVL} variants select the fourth-to-last layer~\cite{chen2024internvl}. The LLaVA series~\cite{liu2023llava, liu2023improvedllava,liu2024llavanext} relies on the penultimate layer. However, these choices are largely heuristics rather than systematic evaluation~\cite{yao2024dense, jiang2023clip, tong2024eyes,cao2024mmfuser,li2026instruction}.

Previous work has shown that ViT layers encode a hierarchy of semantic information—from low-level edge detectors in shallow layers to abstract object representations in deeper layers~\cite{gandelsman2024interpreting, yao2024dense, tong2024eyes}. Yet, how these layer-wise representations affect MLLM performance remains poorly understood. This paper addresses this gap by systematically investigating \emph{which ViT layers provide the most effective visual features for MLLMs}.

We begin by analyzing Layer-wise Representation Similarity (LRS) across CLIP-ViT's hidden states using cosine similarity, revealing three semantically coherent layer groups: shallow (layers 1–12), middle (13–20), and deep (21–24) (Fig.~\ref{fig:cosine_ocr_line_chart}). This categorization provides a foundation for structured layer selection and fusion.

Building upon this foundation, we first systematically assess the efficacy of different deep vision layers. Our analysis reveals that \emph{while the penultimate layer does not universally achieve peak performance in every scenario, it demonstrates consistent superiority across all evaluated model scales} (1.4B, 2.7B, and 7B parameters). This advantage stems from the penultimate layer's unique balance of preserving fine-grained visual details while maintaining strong alignment with textual representations. Notably, the performance gap between the penultimate layer and other deep layers widens as model scale increases. This suggests that \emph{simply using larger LLMs cannot compensate for suboptimal visual feature selection}, underscoring the critical importance of visual layer choice in MLLMs.

Having established the penultimate layer’s strength among deep layers, we ask a more fundamental question: \emph{Can shallower ViT layers offer complementary or even superior information?} Our analysis shows that shallow and middle layers outperform deep layers in approximately one-third of sub-tasks in the MME benchmark~\cite{fu2024mmecomprehensiveevaluationbenchmark} (Fig.~\ref{fig:subtasks4}), particularly in tasks involving fine-grained localization and counting. For instance, layer 18 outperforms the penultimate layer by 20\% on position tasks (Fig.\ref{fig:mme_layer_wise_heatmap}). Similar trends are observed in MMVet~\cite{yu2023mmvetevaluatinglargemultimodal}.  Although shallow layers generally show lower average performance, they still excel on a significant subset of tasks (Fig.\ref{fig:heatmap24layers}). In contrast, deeper layers remain crucial for tasks with high-level semantic demands such as OCR. To assess robustness, we evaluate across three training data scales (665k, 737k, and 1M samples). Despite some fluctuations, our findings consistently demonstrate that \emph{shallow and middle layers carry underutilized yet valuable information}.

Motivated by these insights, we propose a simple yet effective fusion strategy that combines visual features from shallow, middle, and deep layers. Our method uses a single linear projection layer, keeping computational overhead minimal while achieving substantial performance gains. This minimalist approach offers a principled alternative to existing ad-hoc layer selection and fusion methods.
Unlike prior works~\cite{yao2024dense, hong2024cogagent, cao2024mmfuser} that explore hierarchical feature fusion or LLM-aligned selection heuristically, our study provides the first systematic analysis of layer-wise information variation within ViTs, grounded in both intrinsic representation structure and downstream performance.
Our key contributions are summarized as follows:

\begin{figure}[t]
    \begin{subfigure}[b]{0.51\linewidth}  
        \centering
        \includegraphics[width=\linewidth, height=3.5cm, keepaspectratio]{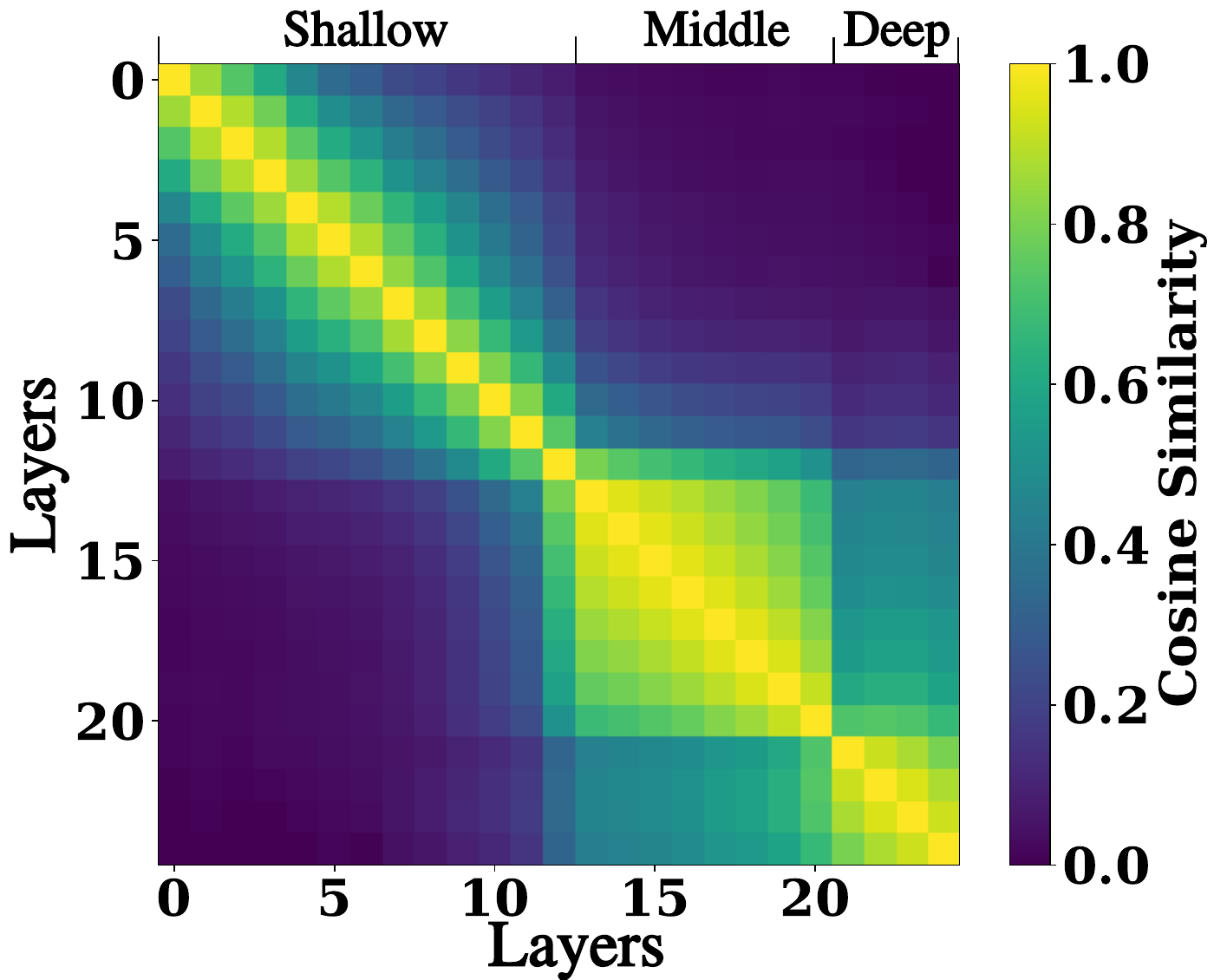}
        \caption{} 
        \label{fig:cosine_similarity}
    \end{subfigure}
    \hspace{0.01\linewidth}  
    \begin{subfigure}[b]{0.46\linewidth}  
        \centering
        \includegraphics[width=\linewidth, height=3.5cm, keepaspectratio]{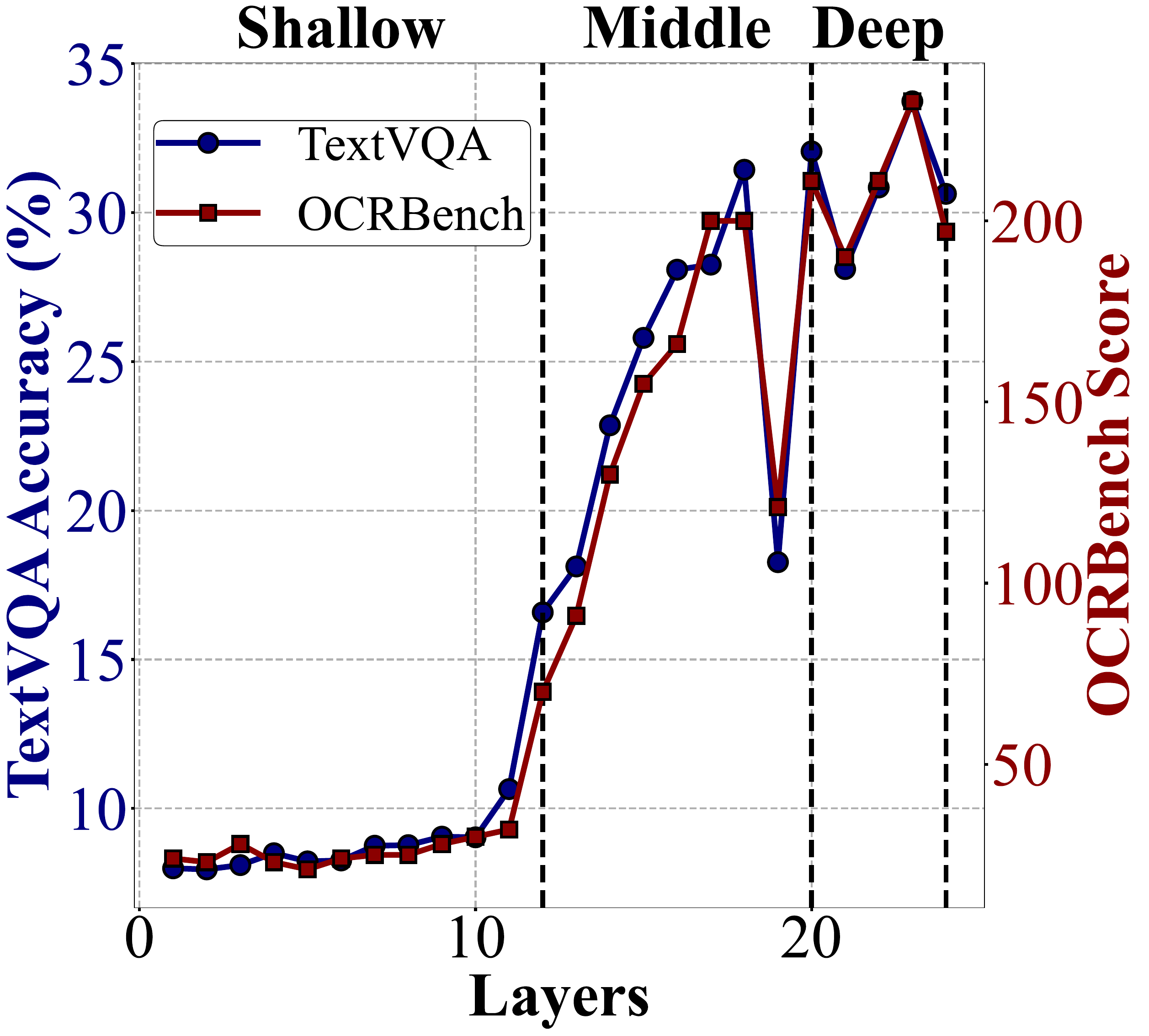}
        \caption{} 
        \label{fig:ocr_line_chart}
    \end{subfigure}
    \caption{\small (a) Average cosine similarity of visual representations across different layers in CLIP-ViT.
(b) Layer-wise performance on OCR tasks. The results highlight three distinct representation regions and their influence on performance.}
    \label{fig:cosine_ocr_line_chart}
\end{figure}

\begin{itemize}
    \item[(1)] We identify three semantically coherent groups of ViT layers (shallow, middle, deep) based on representation similarity. We show that shallow and middle layers, which are often overlooked, can outperform the commonly used deep layers (Sec.~\ref{sec:exp1}).
    \item[(2)] Through extensive experiments across different data sizes and model scales, we confirm the generalization of our findings. Even as gains diminish with scaling, shallow and middle layers continue to exhibit unique strengths over deep layers in certain sub-tasks (Sec.~\ref{sec:exp2}).
    \item[(3)]  We design a linear-layer-based fusion method that integrates features from all three layer groups. It outperforms both specialized fusion designs (e.g., DenseConnector~\cite{yao2024dense}, MMFuser~\cite{cao2024mmfuser}) and standard practices in current MLLMs (e.g., using only the penultimate layer) (Sec.~\ref{sec:exp3}).

 \end{itemize}

\section{Related Work}
\label{sec:relatedwork}
\paragraph{Visual Encoder in Multimodal LLMs}
Serving as the ``eyes'' of MLLMs, the vision encoder sets the upper bound of the model’s perceptual capabilities. CLIP, through image-text contrastive learning effectively aligns visual representation with text space and is widely adopted as the visual encoder in models such as LLaVA \cite{liu2023llava,liu2023improvedllava}, Qwen-VL \cite{wang2024qwen2}, Flamingo \cite{alayrac2022flamingo}, and BLIP \cite{li2023blip}. Other foundational vision models, such as DINOv2 \cite{oquab2023dinov2}, SigLIP \cite{zhai2023sigmoidlosslanguageimage}, ConvNeXT \cite{liu2022convnet2020s}, are also utilized to build MLLMs. In this paper, we select the widely used CLIP-ViT model as the focus of our layer-wise analysis. 
\paragraph{Visual Layer Selection}
Recent studies have explored incorporating shallow visual features within the ViT of multimodal language models, such as DenseConnector~\cite{yao2024dense}, MMFuser~\cite{cao2024mmfuser}. \citet{lin2025multilayervisualfeaturefusion} have further investigated internal fusion strategies by integrating multiple visual layers with language representations, highlighting the critical role of visual layer selection in effective multimodal integration.

Previous methods have largely relied on intuitive, heuristic-based strategies, such as evenly sampling layers. Although some approaches have explored the distinct characteristics of different ViT layers \cite{gandelsman2024interpreting}, the specific roles of layers at different depths in multimodal tasks remain unclear. This study conducts a comprehensive analysis of layer-wise visual representations in MLLMs, aiming to inform the selection of visual layers and guide the design of future visual fusion strategies.

\section{Overall Setup}
\label{sec:preliminaries}
\paragraph{Problem Formulation}

MLLMs typically comprise three core components: a vision encoder, a connector that maps visual features to the language space, and a large language model (LLM)~\cite{lin-etal-2024-preserve}. This architecture empowers MLLMs to handle a diverse array of perception and reasoning tasks across both visual and textual modalities.

Most modern MLLMs adopt a pre-trained CLIP-ViT~\cite{radford2021learning} as their image encoder. A ViT encodes an image into a sequence of token embeddings through a stack of transformer blocks. Each block (or \emph{layer}) progressively refines the visual representations, with earlier layers focusing on low-level spatial details and later layers capturing more abstract, semantic information.

Formally, given an image \( I \), the vision encoder produces a set of layer-wise outputs:
\[
\mathbf{H}^{(1)}, \mathbf{H}^{(2)}, \dots, \mathbf{H}^{(L)} \quad \text{where} \quad \mathbf{H}^{(l)} \in \mathbb{R}^{T \times d}
\]
\( \mathbf{H}^{(l)} \) denotes the embedding at the \( l \)-th layer, \( T \) is the number of tokens, and \( d \) is the dimension.

Despite the availability of rich multi-level features, most MLLMs select a single layer—often the penultimate or final one—to represent the entire image. This practice may overlook complementary signals from shallower layers that encode fine-grained visual details. In this work, we systematically investigate the impact of using different ViT layers for visual input and explore \emph{how selecting appropriate layers can improve MLLM performance across diverse tasks}.

\begin{table*}[!ht]
\centering

\resizebox{1\textwidth}{!}{
\begin{tabular}{cccccccccccccccc}
\toprule
\multirow{2}{*}{Layers} & \multicolumn{5}{c}{General} & \multicolumn{2}{c}{OCR} & \multicolumn{6}{c}{Vision-Centric} & Hallu \\ 
\cmidrule(lr){2-6} \cmidrule(lr){7-8} \cmidrule(lr){9-14} \cmidrule(lr){15-15}
& MME$^P$ & MME$^C$ & MMB & SEEDB & GQA & TVQA & OCRB & CVB & CVB$^{2D}$ & CVB$^{3D}$ & RWD & MMVet & RefCOCO & POPE \\ 
\midrule
\cellcolor{blue!10}1  & 750.1 & 211.4 & 0 & 25.30 & 40.55 & 7.99 & 24 & 40.14 & 34.87 & 45.42 & 38.17 & 9.9 & 5.73 & 70.21 \\
\cellcolor{blue!10}2  & 790.2 & 212.5 & 0.34 & 25.76 & 41.24 & 7.96 & 23 & 40.21 & 34.34 & 46.08 & 37.12 & 10.1 & 5.36 & 71.01 \\
\cellcolor{blue!10}3  & 742.7 & 219.2 & 0.17 & 25.00 & 41.76 & 8.10 & 28 & 42.69 & 37.89 & 47.50 & 36.08 & 10.4 & 6.87 & 72.33 \\
\cellcolor{blue!10}4  & 788.4 & 239.6 & 0 & 25.53 & 42.26 & 8.50 & 23 & 42.69 & 36.71 & 48.67 & 37.25 & 10.2 & 7.61 & 72.34 \\
\cellcolor{blue!10}5  & 813.2 & 220.7 & 0.17 & 25.26 & 42.74 & 8.22 & 21 & 41.30 & 33.69 & 48.92 & 36.86 & 10.9 & 8.25 & 72.91 \\
\cellcolor{blue!10}6  & 838.8 & 227.8 & 0 & 25.23 & 43.16 & 8.26 & 24 & 41.69 & 35.97 & 47.42 & 36.86 & 11.5 & 9.31 & 75.18 \\
\cellcolor{blue!10}7  & 815.6 & 235.7 & 0 & 25.74 & 44.90 & 8.75 & 25 & 43.02 & 37.12 & 48.92 & 37.39 & 10.6 & 11.10 & 75.44 \\
\cellcolor{blue!10}8  & 857.7 & 237.5 & 0 & 25.48 & 46.14 & 8.77 & 25 & 41.43 & 36.85 & 46.00 & 36.99 & 11.2 & 10.79 & 76.20 \\
\cellcolor{blue!10}9  & 889.7 & 232.8 & 0.17 & 27.72 & 47.02 & 9.05 & 28 & 40.53 & 36.23 & 44.83 & 37.12 & 13.0 & 10.06 & 77.84 \\
\cellcolor{blue!10}10 & 903.4 & 228.2 & 0.17 & 26.61 & 48.39 & 9.03 & 30 & 41.8 & 36.19 & 47.42 & 37.39 & 11.4 & 13.50 & 77.46 \\
\cellcolor{blue!10}11 & 935.3 & 224.3 & 0.52 & 26.58 & 49.85 & 10.65 & 32 & 42.51 & 37.27 & 47.75 & 36.86 & 14.2 & 12.14 & 79.24 \\
\cellcolor{blue!10}12 & 980.1 & 232.1 & 0.09 & 26.85 & 50.39 & 16.58 & 70 & 41.81 & 36.7 & 46.92 & 38.56 & 12.6 & 11.20 & 80.63 \\
\midrule

\cellcolor{blue!40}13 & 964.0 & 252.5 & 0.09 & 26.33 & 51.14 & 18.12 & 91 & 41.71 & 35.75 & 47.67 & 37.25 & 11.6 & 12.50 & 81.39 \\
\cellcolor{blue!40}14 & 984.2 & \textbf{265.4} & 0.69 & 34.07 & 51.83 & 22.86 & 130 & 42.69 & 36.12 & 49.25 & 39.08 & 13.8 & 14.37 & 81.97 \\
\cellcolor{blue!40}15 & 1042.8 & 227.5 & 0.17 & 28.98 & 52.89 & 25.79 & 155 & 43.85 & 36.37 & 51.33 & 37.12 & 13.6 & 13.77 & 83.11 \\
\cellcolor{blue!40}16 & 1069.5 & 225.4 & 0 & 27.95 & 52.81 & 28.08 & 166 & 43.26 & 36.61 & 49.92 & 38.04 & 13.7 & 15.41 & 84.32 \\
\cellcolor{blue!40}17 & 1074.8 & 230.4 & 0.26 & 32.81 & 53.86 & 28.25 & 200 & 47.26 & 39.43 & \textbf{55.08} & 39.22 & 15.4 & \textbf{18.49} & 84.46 \\
 
\cellcolor{blue!40}18 & 1088.7 & 237.1 & 29.38 & 52.06 & 54.37 & 31.44 & 200 & \textbf{47.29} & \textbf{41.17} & 53.42 & 39.48 & 14.3 & 17.04 & 84.26 \\
\cellcolor{blue!40}19 & 945.1 & 236.8 & 20.02 & 44.64 & 48.32 & 18.27 & 121 & 45.69 & 37.21 & 54.17 & 35.95 & 13.2 & 18.22 & 81.47 \\
\cellcolor{blue!40}20 & 1118.2 & 232.1 & 26.03 & 51.72 & \textbf{54.83} & 32.05 & 211 & \textbf{47.29} & 40.32 & 54.25 & 38.82 & 16.3 & \textbf{18.49} & 84.76 \\
\midrule
\cellcolor{blue!60}21 & 1041.4 & 212.5 & 0.95 & 35.42 & 49.47 & 28.10 & 190 & 44.37 & 39.32 & 49.42 & 39.87 & 14.5 & 17.09 & 81.91 \\
\cellcolor{blue!60}22 & 1123.6 & 238.9 & 23.28 & 49.60 & 54.52 & 30.84 & 211 & 44.37 & 36.73 & 52.00 & 39.87 & 17.3 & 16.32 & \textbf{84.79} \\
\cellcolor{blue!60}23 & \textbf{1142.7} & 245.0 & \textbf{35.31} & \textbf{52.84} & 54.61 & \textbf{33.73} & \textbf{233} & 44.26 & 38.02 & 50.50 & \textbf{45.36} & \textbf{18.0} & 17.08 & 84.00 \\
\cellcolor{blue!60}24 & 1114.1 & 243.5 & 32.65 & 51.09 & 53.61 & 30.63 & 197 & 46.68 & 39.78 & 53.58 & 43.92 & 16.1 & 17.08 & 83.65 \\
\bottomrule
\end{tabular}
}
\caption{\small Performance across layers $1\text{–}24$. MME$^P$ and MME$^C$ represent the MME perception and cognition tasks respectively. SEEDB, GQA, OCRB and CVB refer to SEEDBench, General QA tasks, OCRBench and CVBench, with CVB$^{2D}$ and CVB$^{3D}$ indicating the 2D/3D subtasks of CVBench, respectively. RWD stands for RealWorldQA. This table provides a detailed analysis of all 24 layers, highlighting that \emph{many optimal performances are found in the middle layers}, which are marked in bold.}
\label{tab:1-24layers_all}
\end{table*}

\paragraph{Partitioning of Visual Representations}
To examine the behavioral patterns of different visual layers, we analyze the relationships between them based on cosine similarities. Inspired by prior findings \cite{sun2024transformer} that LLMs exhibit several distinct representation spaces through such analysis, we similarly identify three significantly different representation spaces within CLIP-ViT. 

As shown in Fig.\ref{fig:cosine_similarity}, three distinct representation spaces emerge among the visual layers (see Appendix~\ref{appendix:A} for computational details). Experiments on OCR and TextVQA (Fig.~\ref{fig:ocr_line_chart}) further indicate that shallow layers contribute little to performance, which rises substantially in the middle layers and peaks in the deep layers. Layers within the same representation space exhibit similar behaviors.

Based on behavioral similarity, we categorize the 24 CLIP-ViT visual layers into three groups: \emph{
shallow layers (1 to 12),
middle layers (13 to 20), and
deep layers (21 to 24)}.

\paragraph{Implementation Details}
We employ CLIP ViT-L/14 (336px)~\cite{radford2021learning} as the visual encoder and 1.4B MobileLLaMA \cite{chu2024mobilevlmv2fasterstronger} as the language model for efficiency analysis, with a one-layer MLP serving as the connector. Training follows a two-phase strategy aligned with LLaVA~\cite{liu2023llava}. AdamW optimizer with a cosine annealing scheduler is used, with learning rates of 1e-3 (phase one) and 2e-5 (phase two), and batch sizes of 256 and 128. Training on four NVIDIA A100 80GB GPUs takes 2 hours for phase one and 8 hours for phase two. We adopt the LLaVA 1.5~\cite{liu2023llava} dataset, comprising 558K image-caption pairs which is carefully filtered from CC3M~\cite{sharma2018conceptual} for pre-training and 665K conversational instances for instruction tuning. Unless explicitly noted, the experimental setup remains the same.
\label{sec:implement_detail}

\paragraph{Evaluation Benchmarks}
\label{sec:benchmarks}

To comprehensively explore and evaluate various visual representations, we classified the benchmarks into four categories following previous work~\cite{tong2024cambrian}: General tasks, OCR tasks, Vision-centric tasks, and Hallucination tasks. 

The \textbf{General tasks} category assesses basic vision-language reasoning abilities, including MME \cite{fu2024mmecomprehensiveevaluationbenchmark} (yes/no questions on attributes like existence and color), MMBench \cite{liu2024mmbenchmultimodalmodelallaround} (multiple-choice across diverse aspects), SEEDBench \cite{li2023seedbenchbenchmarkingmultimodalllms} (spatial and temporal reasoning), and GQA \cite{hudson2019gqanewdatasetrealworld} (complex real-world VQA). The \textbf{OCR category} evaluates a model’s ability to recognize textual content from images, featuring TextVQA \cite{singh2019vqamodelsread} and OCRBench \cite{liu2024ocrbenchhiddenmysteryocr}. The \textbf{Vision-centric category} emphasize fine-grained perception and localization, including CVBench \cite{tong2024cambrian} (evaluating spatial relations and depth), RealWorldQA (real-world QA), MMVet \cite{yu2023mmvetevaluatinglargemultimodal} (general multimodal assessment), and RefCOCO \cite{yu2016modeling} (visual grounding). Finally, the \textbf{Hallucination category} includes POPE \cite{li2023evaluatingobjecthallucinationlarge}, which evaluates whether MLLMs generate false or invented content not grounded in the image.

\begin{figure}[t]
    \centering
    \includegraphics[width=1\linewidth]{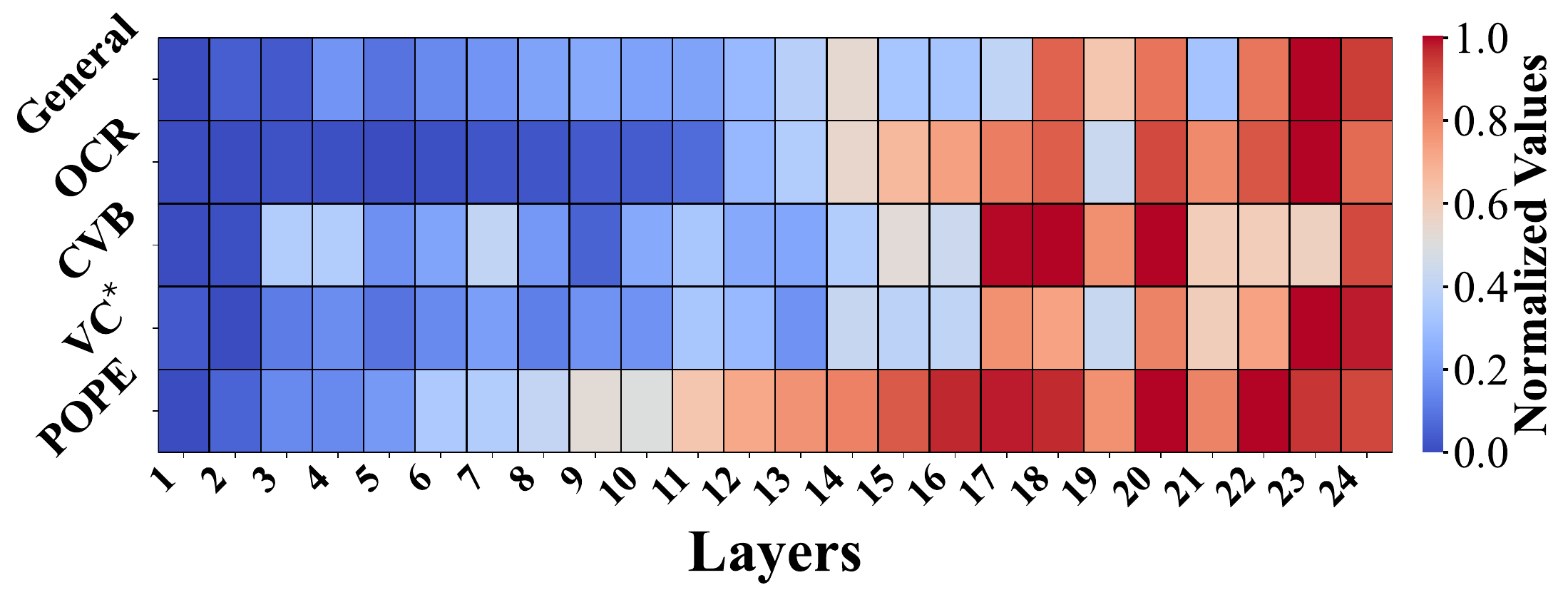}  
    \caption{\small Averaged performance of layers 1 to 24 across various tasks. \(\mathbf{General}\) represents tasks from MME, MMBench, GQA, and SEEDBench. \(\mathbf{OCR}\) includes includes TextVQA and OCRBench. \(\mathbf{CVB}\) corresponds to CVBench, whereas \(\mathbf{VC}^{*}\) includes RefCOCO, RealWorldQA, and MMVet. Results show that the final layer underperforms the penultimate layer, and middle layers sometimes surpass deeper ones.}
    \label{fig:heatmap24layers}
\end{figure}

\begin{figure*}[ht]

    \centering
    \subfloat[MME]{
        \includegraphics[height=2.26cm]{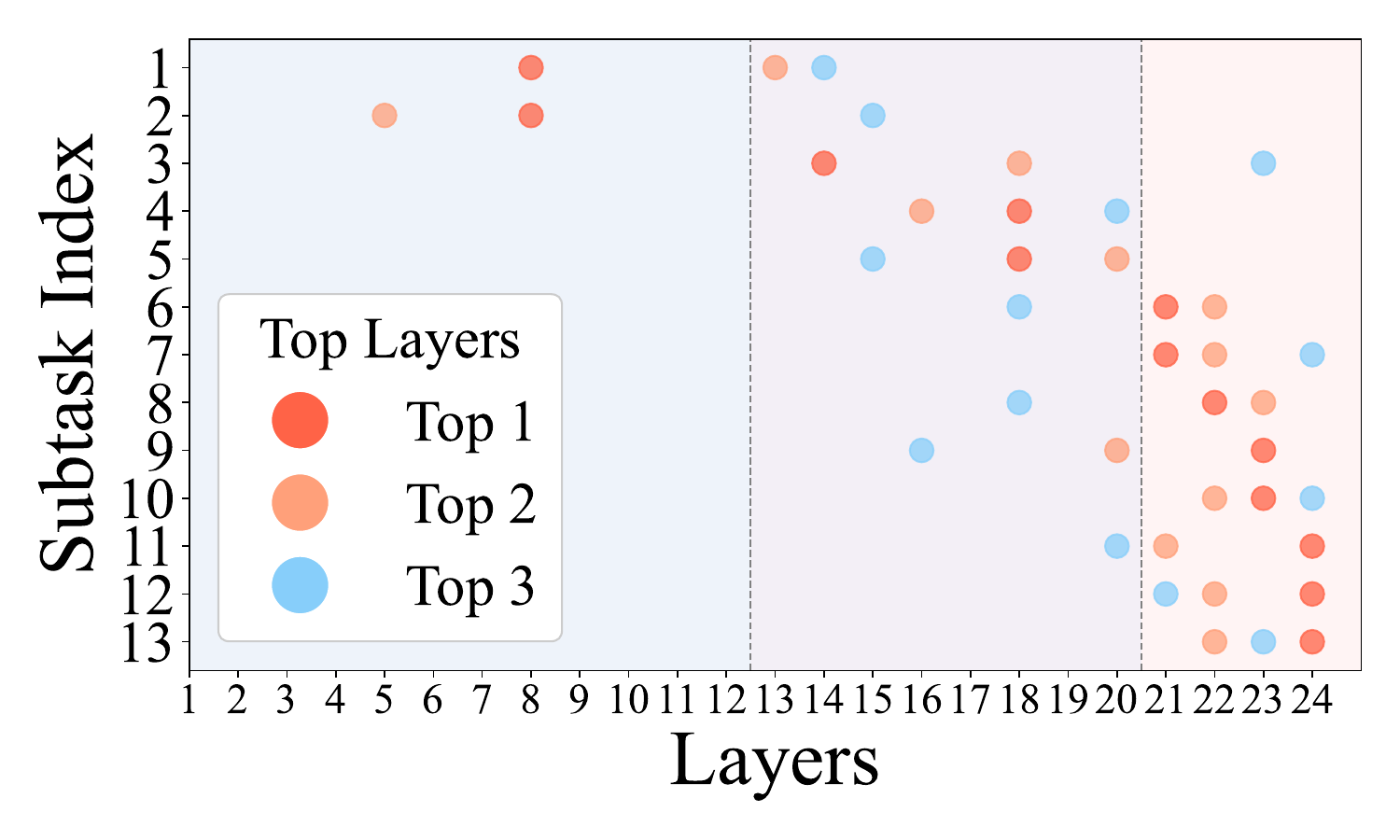}
        \label{fig:mme_scatter}
    } \hspace{-6pt}
    \subfloat[MMVet]{
        \includegraphics[height=2.26cm]{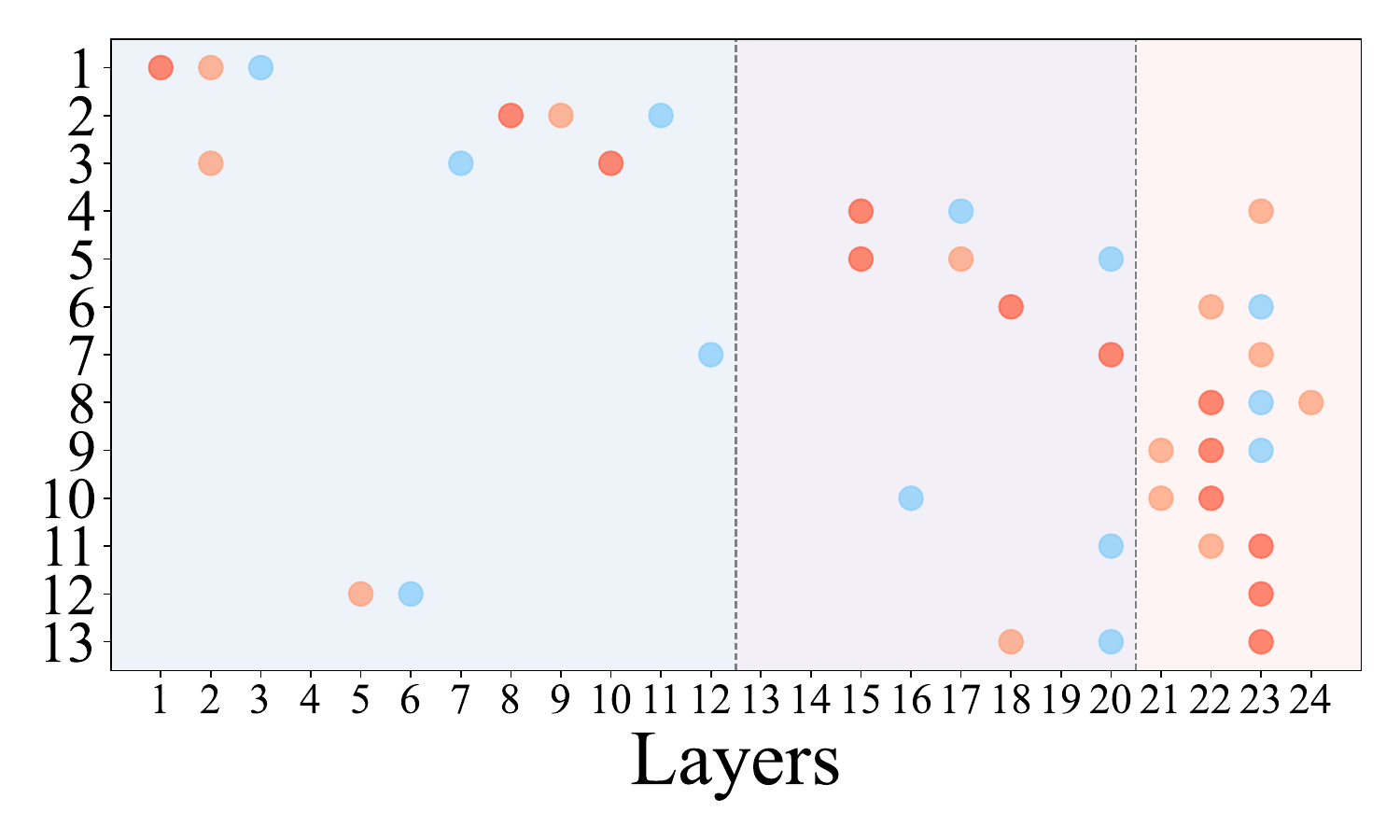}
        \label{fig:mmvet_scatter}
    } \hspace{-6pt}
    \subfloat[MMBench]{
        \includegraphics[height=2.26cm]{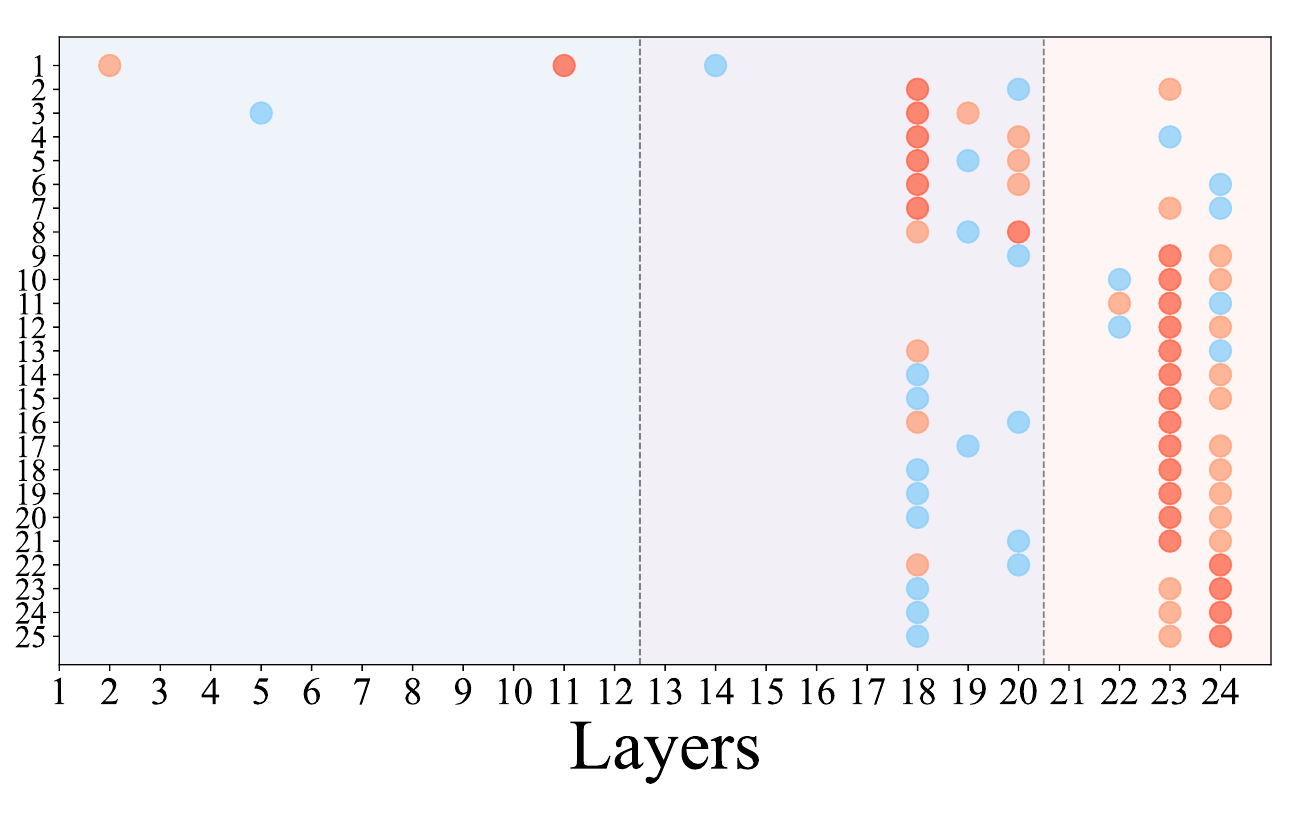}
        \label{fig:mmbench_scatter}
    } \hspace{-6pt}
    \subfloat[SEEDBench]{
        \includegraphics[height=2.26cm]{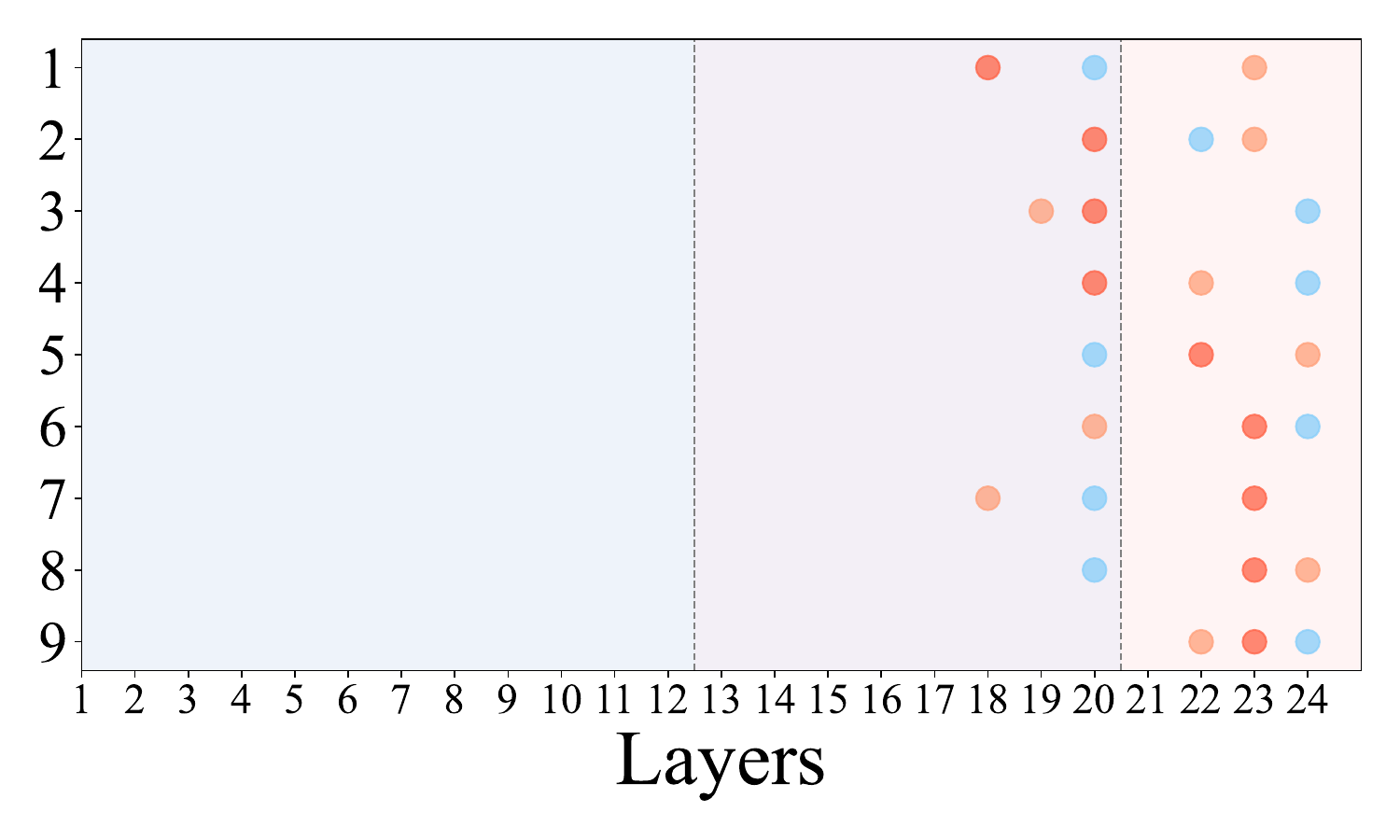}
        \label{fig:seedbench_scatter}
    }
    \vspace{-6pt}
    \caption{\small Layer-wise performance distribution across four benchmarks: \textbf{(a) MME}, \textbf{(b) MMVet}, \textbf{(c) MMBench}, and \textbf{(d) SEEDBench}. The x-axis corresponds to layer indices and the y-axis indicates the sub-tasks index (see Tab.~\ref{appendix:subtasks2index} for details). Top-performing layers for each sub-task are highlighted with color-coded markers: \textcolor[rgb]{0.980, 0.412, 0.306}{$\bullet$} (1st place), \textcolor[rgb]{0.984, 0.639, 0.494}{$\bullet$} (2nd place), and \textcolor[rgb]{0.557, 0.800, 0.973}{$\bullet$} (3rd place). }
    \label{fig:subtasks4}
\end{figure*}
\section{Experiment: Layer-wise Exploration}
\label{sec:exp1}

Previous studies have primarily used techniques such as linear probing and attention head decomposition to analyze CLIP-ViT representations~\cite{gandelsman2024interpreting}. While these methods reveal what types of information are present in different ViT layers, they do not assess whether such information can be effectively utilized by MLLMs. The mere presence of information in a particular layer does not guarantee its usefulness when integrated into an MLLM. In contrast, our work goes beyond probing for representational content—we systematically evaluate how each ViT layer contributes to downstream MLLM performance. To this end, we conduct a layerwise exploration by individually connecting each visual layer to the language model, training the corresponding MLLM, and benchmarking its task performance. The layerwise performance is shown in Tab.~\ref{tab:1-24layers_all} and Fig.~\ref{fig:heatmap24layers}.

\subsection{Deep-to-Deep Layer Comparison}
\label{sec:exp1-2}

A common practice is to use deep layers from ViT as input to the MLLM. In this section, we investigate the effectiveness of this approach.

\paragraph{The final layer is not the optimal choice:} As shown in Tab.~\ref{tab:1-24layers_all} and Fig.~\ref{fig:heatmap24layers}, the final layer does \emph{NOT} perform the best on any benchmark. For general tasks, a noticeable performance drop is observed at the final layer, with OCR tasks exhibiting particularly severe degradation. A similar trend is evident in POPE. However, vision-centric tasks partially show this decline. Overall, these results indicate that the final layer is \emph{NOT} the optimal choice for representation across tasks.

The underlying reason might lie in the CLIP model’s training mechanism, where supervision primarily focuses on aligning the final layer $[\text{CLS}]$ token with text embeddings. The $[\text{CLS}]$ token in the final layer is optimized by CLIP’s contrastive loss, making it highly specialized for the image-text matching task. However, this optimization process can significantly suppress local image features in the final layer, which can harm performance when finer-grained details are required.

\paragraph{Penultimate layer as the optimal choice.} As shown in Fig.~\ref{fig:heatmap24layers}, the penultimate layer consistently achieves the best performance across tasks. Notably, it outperforms other deep layers on General, OCR, and vision-centric tasks. This superiority stems from its ability to retain rich visual information while maintaining strong text alignment, which is second only to the final layer. Such a balance offers an optimal trade-off between visual expressiveness and semantic alignment, making it particularly well-suited for multimodal tasks.

\subsection{Deep-to-Shallower Layer Comparison}

\label{sec:layer-wise-analysis}
Afterwards, we investigate the effects of using shallow and middle layers in MLLMs. We highlight the following key observations.

\paragraph{Deep layers are essential for OCR.} 
As shown in Fig.~\ref{fig:cosine_ocr_line_chart}, shallow layers provide negligible text information for the LLM. A clear boundary exists between the shallow and middle layers, with layer 12 marking the transition point. Layers before this point fail to contribute meaningfully to text processing, while a sharp performance gain occurs immediately afterward. This might be attributed to two essential requirements for OCR tasks:
\begin{enumerate}
    \item \emph{Rich fine-grained visual features in visual representation}: In OCR tasks, a strong perception of details is often required. Therefore, these fine-grained details must be embedded in the representation with sufficiently strong signals to be effectively utilized by the LLM.
    \item \emph{Well textually aligned visual features}: Despite containing rich visual details, shallow layers lack intrinsic alignment with textual representations, limiting their usefulness for OCR tasks. Fig.~\ref{fig:cosine_similarity} confirms this with notably low cosine similarity between shallow and deep layers. This discrepancy poses a challenge for the connector, which can only align features originating from the deep (text) space or adjacent middle layers.
\end{enumerate}

\paragraph{Limited impact of representation quality on cognitive tasks.}
Cognitive tasks, such as ``Code Reasoning'' and ``Numerical Calculation'', require both perception and high-level reasoning capabilities. Interestingly, we observe that even the shallow layers, which generally yield lower quality visual representations can rival or even outperform deeper layers in tasks under the MME-Cognitive. layer 3 achieves superior performance on ``Code Reasoning'', ``Numerical Calculation', and ``Text Recognition'' compared to both middle and deep layers. (see Tab.\ref{tab:seedbench-2.7-7b} and Tab.\ref{tab:MME-737k-1M} in Appendix) \textit{These findings suggest that, for cognitive tasks, visual feature quality is not the primary limiting factor. }

\paragraph{Potential of middle layers} We first conduct an investigation into the middle representation spaces. The performance of these two spaces, as shown in Tab.~\ref{tab:1-24layers_all}, several key insights emerge from the results:

\textbf{(1)} \emph{The middle layer has the potential to perform best:} Although the middle layer’s information has not been fully processed, it still achieves the best performance on one-third of the benchmarks. Specifically, compared to the penultimate layer, layer 14 achieves a 20-point higher score on MME-Cognitive, layer 18 outperforms by 3\% on CVBench, layer 17 surpasses by 1.4\% on RefCOCO, and layer 20 exceeds by 0.2\% on GQA.

\textbf{(2)} \emph{The middle layers generally perform better on vision-centric tasks:} Fig.~\ref{fig:mme_scatter} illustrates the performance of different layers across subtasks in the MME dataset, showing that position and existence tasks benefit more from middle layer representations. As depicted in Fig.~\ref{fig:mmvet_scatter}, the penultimate layer achieves top performance in only three out of eleven subtasks, whereas the shallow and middle layers yield optimal results in seven. Similarly, in Fig.~\ref{fig:mmbench_scatter}, one-third of the best performing results, such as those in spatial relations, physical relations, and cross fine-grained perception originate from the shallow and middle layers. A comparable trend is observed in SEEDBench (Fig.~\ref{fig:seedbench_scatter}), where middle layers produce optimal results in nearly half of the subtasks, including Instance Attribute, Instance Location, Instance Interaction, and Text Recognition.

\paragraph{The hallucination problem is more pronounced in shallow layers but is effectively mitigated in the middle layers.}  
As shown in Tab.~\ref{tab:1-24layers_all}, POPE results indicate that hallucination issues are most prominent in the shallow representation space, with minimal variation between the middle and deep layers. Notably, in the middle representation space, half of the layers outperform the penultimate layer on the POPE. This phenomenon likely stems from the fact that the challenge of this task lies more in visual perception than in semantic comprehension. In Sec.\ref{sec:exp2-2}, we provide a detailed analysis showing that further experiments with larger LLMs consistently support this finding.

\begin{table*}[t]
\vspace{-20pt}
\centering
\renewcommand{\arraystretch}{1.15}
\setlength{\tabcolsep}{4pt}
\resizebox{0.85\textwidth}{!}{
\setlength{\tabcolsep}{4pt}
\begin{tabular}{cccccccccccccc}
\toprule
\multirow{2}{*}{\textbf{Data Scale}} & \multirow{2}{*}{\textbf{Layers}} & \multicolumn{5}{c}{\textbf{General}} & \multicolumn{2}{c}{\textbf{OCR}} & \multicolumn{4}{c}{\textbf{Vision-Centric}} & \textbf{Hallu} \\
\cmidrule(lr){3-7} \cmidrule(lr){8-9} \cmidrule(lr){10-13} \cmidrule(lr){14-14}
 &  & MME$^{P}$ & MME$^{C}$ & MMB & SEEDB & GQA & TVQA & OCRB & CVB & CVB$^{2D}$ & CVB$^{3D}$ & RWQA & POPE \\
\hline
\multirow{4}{*}{\textbf{665k}} 
& \textbf{3} & 742.7 & 219.3 & 0.17 & 25.00 & 41.76 & 8.10 & 28 & 42.69 & 37.89 & 47.50 & 36.08 & 72.33 \\
& \textbf{18} & 1088.8 & 237.1 & 29.38 & 52.06 & 54.37 & 31.44 & 200 & \textbf{47.29} & \textbf{41.17} & 53.42 & 39.48 & \textbf{84.26} \\
& \textbf{23} & \textbf{1142.8} & \textbf{245.0} & \textbf{35.31} & \textbf{52.84} & \textbf{54.61} & \textbf{33.73} & \textbf{233} & 44.26 & 38.02 & 50.50 & \textbf{45.36} & 84.00 \\
& \textbf{24} & 1114.1 & 243.6 & 32.65 & 51.09 & 53.61 & 30.63 & 197 & 46.68 & 39.78 & \textbf{53.58} & 43.92 & 83.65 \\
\hline
\multirow{4}{*}{\textbf{737k}} 
& \textbf{3} & 845.9 & 225.4 & 0.26 & 26.33 & 44.12 & 8.34 & 27 & 42.36 & 35.81 & 48.92 & 37.52 & 74.98 \\
& \textbf{18} & 1093.1 & 226.4 & 43.04 & \textbf{56.33} & \textbf{56.98} & 35.98 & \textbf{270} & \textbf{48.87} & 46.57 & \textbf{51.17} & 43.40 & 86.18 \\
& \textbf{23} & \textbf{1163.7} & 230.0 & \textbf{48.37} & 55.55 & 56.77 & \textbf{36.41} & 265 & 48.09 & \textbf{47.59} & 48.58 & 41.83 & \textbf{86.22} \\
    & \textbf{24} & 1121.7 & \textbf{258.6} & 46.05 & 55.43 & 56.34 & 36.09 & 255 & 43.63 & 38.01 & 49.25 & \textbf{44.44} & 85.09 \\
    \hline
    \multirow{4}{*}{\textbf{1M}} 
& \textbf{3} & 871.8 & 215.4 & 13.40 & 40.21 & 45.67 & 8.03 & 26 & 43.04 & 37.74 & 48.33 & 39.08 & 74.08 \\
& \textbf{18} & 1145.9 & 213.2 & 42.44 & 56.72 & 57.74 & 35.68 & \textbf{267} & \textbf{54.08} & \textbf{56.33} & 51.83 & 43.14 & 84.06 \\
& \textbf{23} & \textbf{1214.4} & \textbf{249.3} & \textbf{52.92} & \textbf{58.58} & \textbf{57.91} & \textbf{37.24} & 263 & 53.48 & 53.96 & \textbf{53.00} & 43.27 & 84.03 \\
& \textbf{24} & 1192.0 & 245.0 & 47.34 & 57.62 & 57.21 & 36.45 & 264 & 47.88 & 42.84 & 52.92 & \textbf{44.97} & \textbf{84.58} \\
\hline
\end{tabular}
}
\caption{\small Performance comparison of visual representations across different data scales, demonstrating the consistency of our key findings. Even as gains diminish with scaling, middle layers continue to exhibit unique strengths over deep layers in OCRBench, SEEDBench, GQA and CVBench.}
\label{tab:data-scale-compare}
\end{table*}
\begin{figure}
    \centering
    \includegraphics[width=\linewidth]{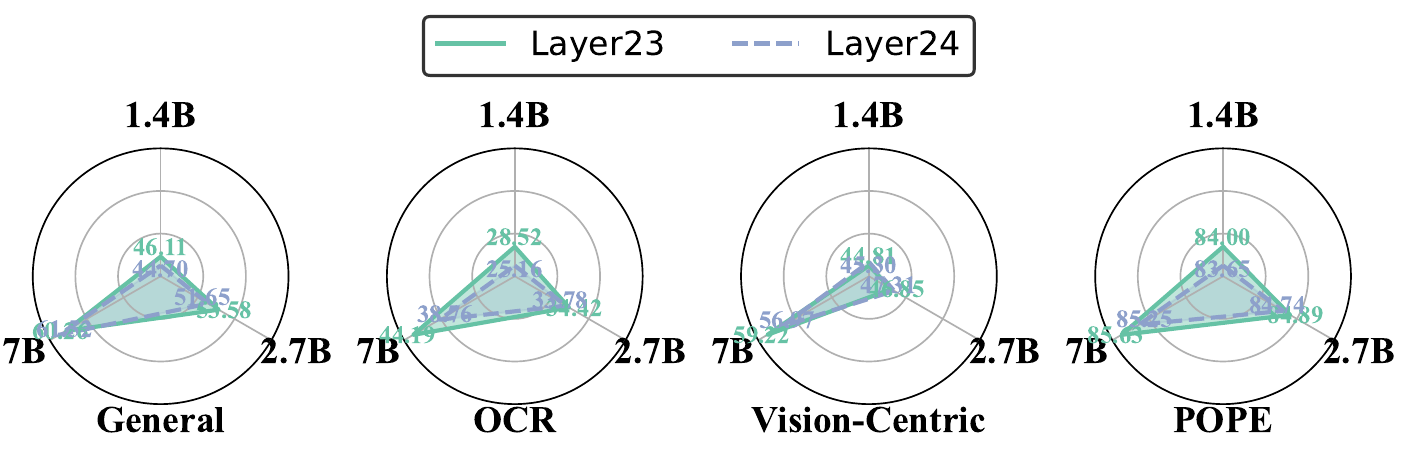}
    \caption{\small Radar charts comparing the performance of Layers 23 and 24 across four different tasks under three LLM scales: 1.4B, 2.7B, and 7B. The results consistently show that \emph{the penultimate layer outperforms the final layer in all tasks}. This trend remains stable across different model scales.}
    \label{fig:miniradar}
\end{figure}

\begin{figure}[t]
    \centering
    \includegraphics[width=0.85\linewidth]{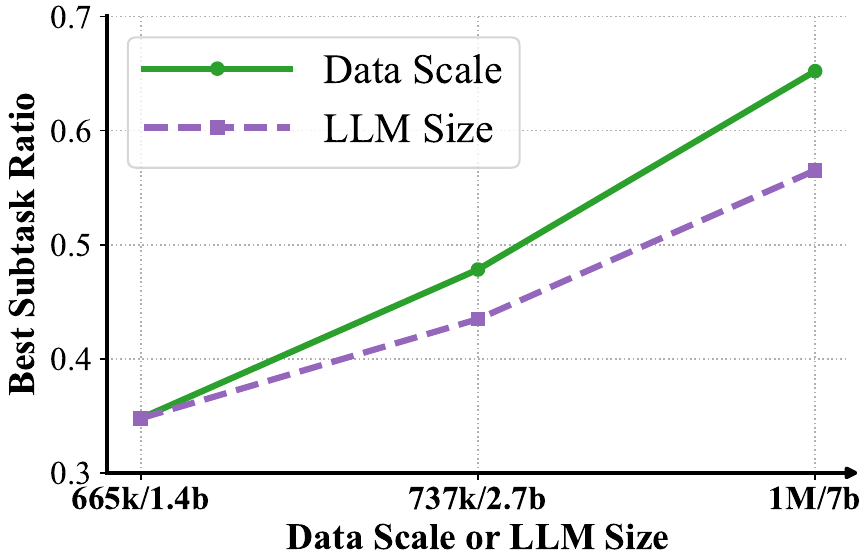}
    \caption{\small Proportion of subtasks achieving their best performance at the penultimate layer on MME and SEEDBench, demonstrating a clear upward trend.}
    \label{fig:scale-line-chart}
\end{figure}
\section{Effect of Data and Model Scale}
\label{sec:exp2}
To further assess the generality of our findings, we extend our experiments to larger model scales and training datasets in this section and analyze the resulting performance trends.

\subsection{Settings}

\label{sec:exp2setting}
\paragraph{More training Data.} We investigate the impact of training data scales using the same training procedure detailed in Sec.~\ref{sec:implement_detail}. In the first stage, we use the LLaVA 558k dataset~\cite{liu2023llava}. For the second stage, we evaluate three dataset configurations: \textbf{(1)} LLaVA 665K, \textbf{(2)} Cambrian-1 737K~\cite{tong2024cambrian}, an expansion of the 665K dataset with additional OCR data, and \textbf{(3)} a custom 1M dataset that builds on the 737K dataset by incorporating data specifically curated for vision-centric tasks. The dataset composition can be found in Appendix~\ref{appendix:exp_details}.
\paragraph{Scaling LLM sizes.} Building on the original 1.4B experiments, we extend our study to include MobileLLaMA~\cite{chu2024mobilevlmv2fasterstronger} 2.7B and Vicuna v1.5 7B. These LLMs are selected due to their similar architectures, making them well-suited for investigating the impact of different LLM sizes.

Due to computational constraints, we do not conduct a full layer-wise analysis across different data scales. Instead, leveraging insights from Sec~\ref{sec:layer-wise-analysis}, we select representative layers from the shallow (layer 3), middle (layer 18), and deep (layers 23 and 24) representation spaces to examine how variations in data scale affect model performance.

\subsection{Deep-to-Deep Layer Comparison }

As shown in Tab.~\ref{tab:data-scale-compare}, our findings indicate that the key conclusions in Sec.~\ref{sec:layer-wise-analysis} remain valid across different training data scales. As data scales up, we uncover the following key insights:
\paragraph{The penultimate layer remains the optimal choice in deep space regardless of LLM size} As shown in Fig.~\ref{fig:miniradar}, the penultimate layer consistently outperforms the final layer across LLMs of 1.4B, 2.7B, and 7B, reinforcing our findings. This indicates that CLIP-ViT’s final-layer visual degradation, driven by its training paradigm, cannot be offset by a stronger LLM. 

Furthermore, increasing LLM size does not yield significant improvements in POPE performance, indicating that hallucination bottlenecks in MLLMs stem primarily from the quality of the visual representation. In contrast, vision-centric tasks benefit more from scaling LLM size, indicating that even for tasks grounded in visual understanding, strong perception alone does not suffice, as robust reasoning capabilities remain essential.

\paragraph{The more data and the larger the model, the more the deep layers benefit.}
As both the training data scale and LLM size increase, a clear trend emerges in Fig.~\ref{fig:scale-line-chart}: the proportion of subtasks where the penultimate layer achieves the best performance consistently grows. 

Since the visual encoder remains frozen during training, this suggests that compared to the middle layers, fine-grained information is less explicitly preserved in the deep layers. In other words, fine-grained details in the middle layers are more readily utilized by the LLM, whereas those in the deep layers are harder to extract, requiring larger amounts of data to activate the LLM to effectively capture these fine-grained features.

\subsection{Deep-to-Shallower Layer Comparison}
\label{sec:exp2-2}
We observe that while conclusions from small models may not fully generalize to larger ones, the main findings still hold, as detailed below:
\paragraph{The potential of shallower layers persists across model and data scales}
The consistency of the previous conclusion is validated under larger training data and increased LLM size. As detailed in Appendix~\ref{tab:MME-737k-1M} and~\ref{tab:seedbench-737k-1M}, under the 2.7B model, the penultimate layer fails to outperform shallower layers on several MME subtasks, such as Count, Position, and Existence. Similar patterns emerge in SEEDBench and persist with the 7B model, where shallower layers (e.g., Layer 18) achieve better results on tasks like Spatial Relation. 
On the 665K dataset, layer 18 outperforms the penultimate layer by up to 3\% on CVBench and maintains a slight advantage on OCR and vision-centric tasks. This trend persists on the 737K dataset, where layer 18 continues to lead on SEEDBench and GQA. Although the performance gap narrows on the 1M dataset, the middle layer still surpasses the deep layer on both OCRBench and CVBench.

\paragraph{Limited gains on OCR tasks despite data and LLM scaling.} Layer 3 exhibits comparable performance on OCR tasks both before and after incorporating OCR-specific training data, and across models of different scales (2.7B vs. 7B). This suggests that \emph{increasing task-specific data alone cannot overcome the inherent limitations of shallow representations}. The lack of improvement can be attributed to their poor alignment with the textural feature space required for OCR understanding. In contrast, the middle layers, though only partially aligned, still exhibit performance gains under additional supervision.

\begin{table*}[t]
    \centering
    \renewcommand\tabcolsep{2.4pt}
    \resizebox{0.8\textwidth}{!}{ 
    \begin{tabular}{lccccccccccccc}
        \toprule
        \multirow{2}{*}{\textbf{Models}} & \multicolumn{5}{c}{\textbf{General}} & \multicolumn{2}{c}{\textbf{OCR}} & \multicolumn{4}{c}{\textbf{Vision-Centric}} & \multicolumn{1}{c}{\textbf{Hallu}} \\
        \cmidrule(lr){2-6} \cmidrule(lr){7-8} \cmidrule(lr){9-12} \cmidrule(lr){13-13}
        & MME$^P$ & MME$^C$ & MMB & SEEDB & GQA & TextVQA & OCRB & CVB & CVB$^{2D}$ & CVB$^{3D}$ & RWQA & POPE &  \textbf{Win} \\
        \midrule
        \rowcolor{gray!15}{Baseline(${23}$)}   & 1142.8  & 245.0  & 35.31  & 52.84  & 52.84  & 33.73  & 233  & 44.26  & \textbf{38.02}  & 50.50  & 45.36  & 84.00 &\textcolor{PineGreen}{\textbf{9/10}} \\
        \midrule
        {DenseConnector}  & 1145.0  & \textbf{253.2}  & 47.85  & 57.16  & 56.92  & 37.54  & 257  & 45.60  & 35.83  & \textbf{54.92}  & 45.10  & \textbf{84.95}  &\textcolor{PineGreen}{\textbf{7/10}} \\
        {MMFuser}  & 1149.5  & 238.9  & \textbf{49.65}  & 56.21  & 56.59  & 35.43  & 245  & \textbf{45.70}  & 36.89  & 54.50  & 44.83  & 84.53 &\textcolor{PineGreen}{\textbf{8/10}} \\
        \rowcolor{gray!15}\({\text{Ours}}^{*}\)  & \textbf{1157.2}  & 236.1  & 49.22  & \textbf{57.23}  & \textbf{57.35}  & \textbf{37.70}  & \textbf{265}  & 44.56  & 36.53  & 52.58  & \textbf{45.75}  & 84.82 & - \\
        \bottomrule
    \end{tabular}
    }
    \caption{\small Study on different layer fusion strategies. ``Ours'' represents \(\mathcal{L}_5 = {23,18,3}\). ``Win'' denotes the proportion of datasets where our method achieves superior performance. Our method outperforms DC and MMFuser on 7 and 8 out of 10 benchmarks.}
    \vspace{-5pt}
    \label{tab:exp3-2_table_compare}
\end{table*}
\section{Visual Feature Fusion}
\label{sec:exp3}

Building on the above-mentioned findings that deeper layers are not universally optimal and that shallower layers offer valuable complementary information, we explore the most effective way to enhance visual representation by combining visual features from multiple layers. To be specific, we employ a simple fusion strategy to merge features from different layers and conduct a preliminary study on various layer combinations, aiming to highlight the potential benefits of layer fusion. 

\subsection{Method}
\label{sec:exp3setting}
The equation below shows the simplest visual feature fusion mechanism, 
\begin{equation}
f = \operatorname{Concat}\left(\mathcal{H}^{(i)} \mid i \in \mathcal{L} \right)
\end{equation}
where \( \mathcal{L} \) denotes the set of selected layers, and each \( \mathcal{H}^{(i)} \) represents the feature representation extracted from layer \( l \) with a dimension of \( N \times D \). Here, \( N \) is the number of visual tokens, and \( D \) is the feature dimensionality of each token. The concatenation function \( \operatorname{Concat}(\cdot) \) merges these representations along the feature dimension, producing an output \( f \) of size \( N \times (D \times |\mathcal{L}|) \), where \( |\mathcal{L}| \) denotes the number of concatenated layers. The resulting \( f \) is then fed into the Connector. Subsequently, we explore various layer combinations.

\subsection{Exp-I: Ablation of Fusion Layer Selection}
\label{sec:ablationstudy}

We select representative layers from shallow, middle, and deep layers for the fusion ablation study to preliminarily explore the effect of different representation spaces in fusion methods. We systematically construct different layer combinations $\mathcal{L}$ to analyze their impact on fusion performance.

\begin{figure}[t]
    \centering
    \includegraphics[width=0.9\linewidth]{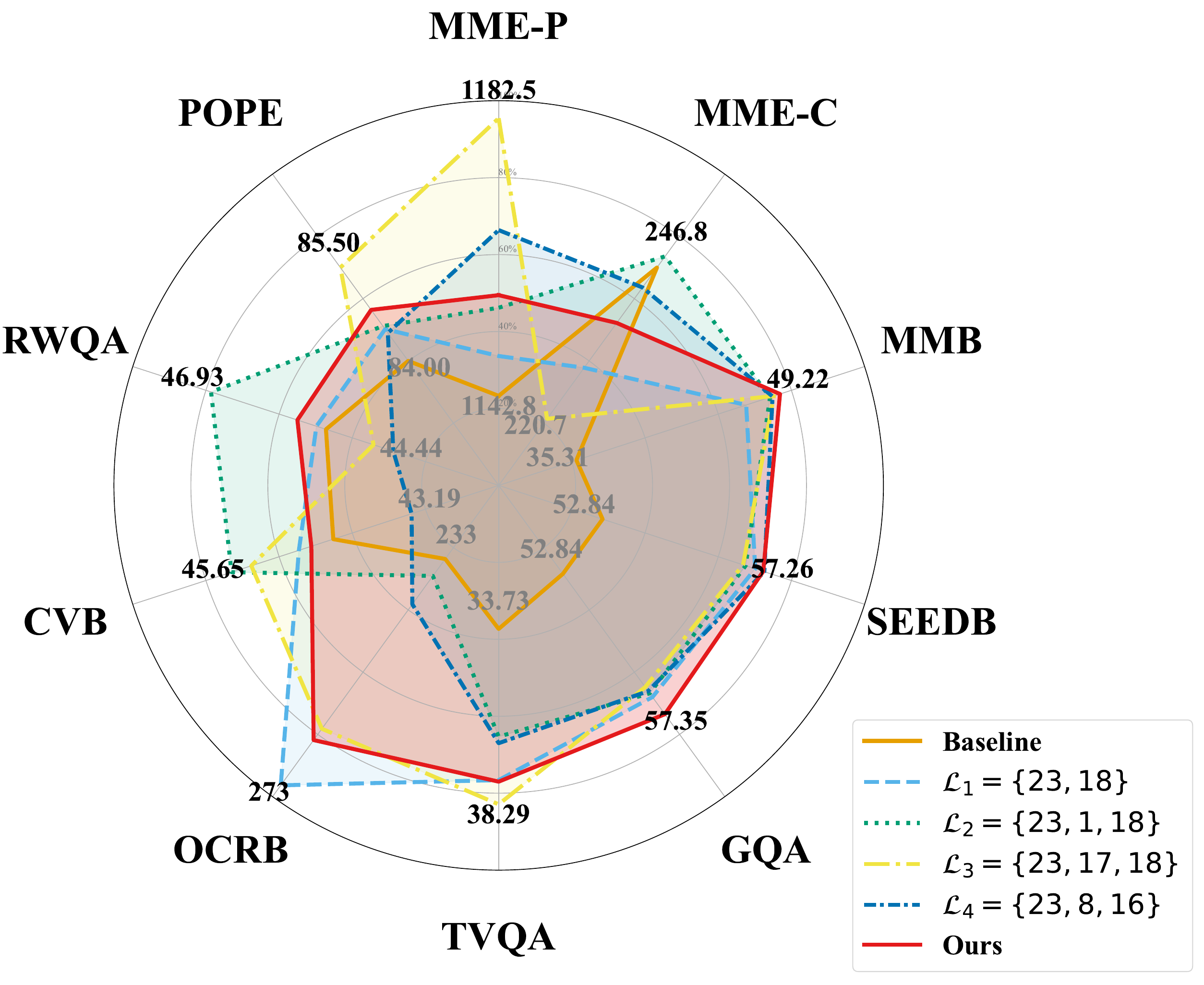}
    \caption{\small Performance comparison of different layer fusion combinations on four tasks. $\mathcal{L}_1$–$\mathcal{L}_4$ denote representative strategies for layer selection. ``Ours'' is $\mathcal{L=}$\{23, 3, 18\}. }

    \label{spider}
\end{figure}

\paragraph{Multiple stages bring generalization:} Fig.~\ref{spider} illustrates six different configurations, ranging from using only the end stage to incorporating all stages. Compared to two-layer fusion (\(\mathcal{L}_1\)) and three-layer fusion covering two stages (\(\mathcal{L}_{3}\)), three-layer fusion (\(\mathcal{L}_{2,3}\)) that spans all stages (\(\mathcal{L}_{2}\) and Ours \{23, 3, 18\}) leads to more consistent performance improvements. Notably, \(\mathcal{L}_{2}\) performs worse than Ours \{23, 3, 18\} on OCR tasks. This is likely because the first layer feature is too raw, making them less suitable for extracting the low-level visual cues required in OCR. By incorporating more stable representations from multiple stages, models can achieve the most robust performance across tasks.

\subsection{Exp-II: Fusion Method Comparison}
\label{sec:exp3-2}

To investigate the impact of different layer fusion strategies, we compare our selected layers \{23, 3, 18\} and a simple concatenation strategy, against other carefully designed fusion methods.

\paragraph{Less is more:} We compare two state-of-the-art ad-hoc methods, DenseConnector (DC) and MMFuser with our method. As shown in Tab.~\ref{tab:exp3-2_table_compare}, \emph{our method outperforms DC and MMFuser on 7 and 8 across 10 widely used benchmarks}, respectively. These results suggest that complex fusion strategies may be unnecessary, as the simplest concatenation already meets the performance requirements.

\section{Conclusion}
This study presents a comprehensive layer-wise analysis, revealing that shallow and middle representation spaces can surpass the performance of deep layers. Evaluations across diverse data and model scales further substantiate our findings. Furthermore, we introduce a straightforward yet highly effective fusion strategy for visual feature integration, delivering substantial improvements over the baseline. Our findings offer a foundation for advancing future research in fusion methodologies.
\section*{Limitations}

Due to the high computational cost of layer-wise analysis, we adopt a linear probing-inspired strategy: most experiments are conducted on the 1.4B model, with selective validation on the 2.7B and 7B variants. However, our study does not extend to larger-scale LLMs. In terms of visual encoders, we focus exclusively on CLIP-ViT-L/14, given its widespread adoption, and leave the exploration of alternative backbones to future work.

Moreover, vision–language fusion strategies can be broadly classified into internal and external methods. Our analysis focuses only on external fusion approaches and does not directly compare them with internal alternatives. We ensure consistent experimental conditions across all settings, which enables a fair assessment of visual representation quality. However, our current design does not explore how different connector architectures may affect performance. We regard this as a valuable direction for future research.

\section*{Acknowledgement}
We are grateful to EIT and IDT High Performance Computing Center for providing computational resources that greatly facilitated this research. This work was financially supported by the 2035 Key Research and Development Program of Ningbo City (Grant No. 2025Z034), the Pioneer and Leading Goose R\&D Pr
ogram of Zhejiang (Grant No. 2024C01110), and the National Natural Science Foundation of China (Grant No. 62472385).

\newpage
\bibliography{main}
\newpage

\setcounter{page}{1}
\appendix
\section*{Appendix}
\label{sec:appendix}

We provide some additional information as supplementary material. This material is divided into four sections:
\begin{itemize}
    \item A detailed analysis and formal computation process of visual representations is provided in Appendix~\ref{appendix:A}.
    \item Experimental details covering dataset composition, evaluation protocols, visualization of fusion performance, and analyses on model factors such as LLM choice, data scale, layer selection, and feature fusion are provided in Appendix~\ref{appendix:exp_details}.
    \item We provide additional details of our reasoning and rebuttal explanations in Appendix~\ref{appendix:C}.
    \item Ethics statement is provided in Appendix~\ref{Appendix:D}.
\end{itemize}

\begin{figure*}[htbp]
    \centering
    \includegraphics[width=\linewidth]{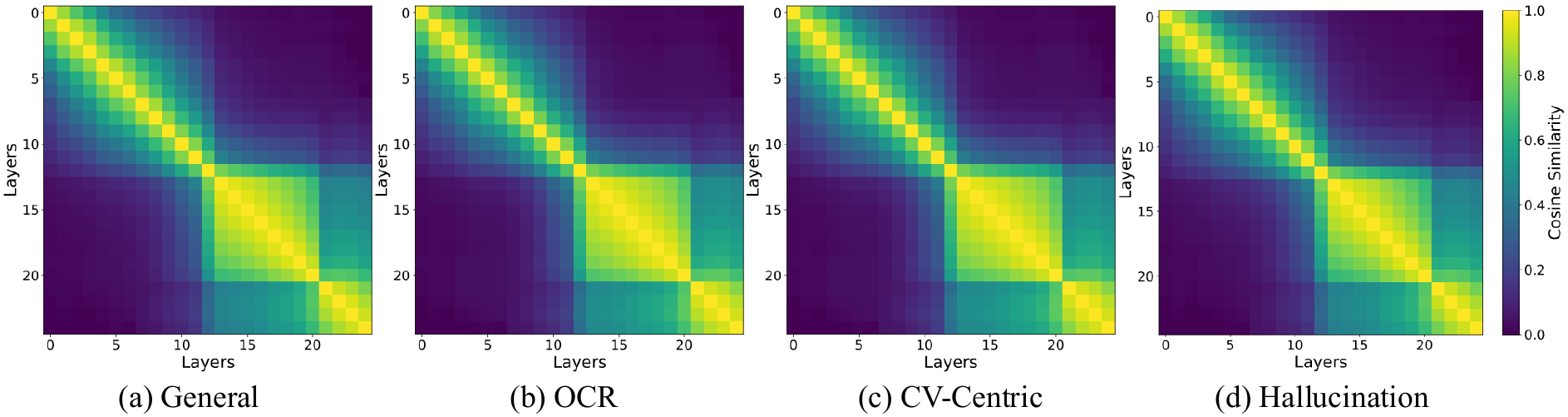}
    \caption{\small A visualization of the average cosine similarity of visual representations across different layers in CLIP-ViT for four tasks, namely General, OCR, CV-Centric, and Hallucination. Values closer to 1 indicate greater similarity.}
    \label{fig:cosine_similarity_clip_4}
\end{figure*}

\section{Visual Representations}
\label{appendix:A}

\subsection{Formal Computation of LRS}
To examine the behavioral patterns of different visual layers, we introduce \textbf{Layer-wise Representation Similarity (LRS)} to analyze the relationships between various visual layers.

Specifically, we compute the cosine similarity between hidden states of ViT layers, where values closer to \(1\) indicate higher similarity. The formalized division process is detailed in the following equation:

We define the hidden state matrix \( H \) of CLIP-ViT as:
\begin{equation*}
H \in \mathbb{R}^{L \times d}
\end{equation*}
where \( L \) is the number of layers, and each row \( H_i \in \mathbb{R}^{d} \) represents the hidden state of the \( i \)-th layer. The cosine similarity matrix \( S \) is computed as:

\begin{equation*}
S = \left| \frac{H H^T}{\|H\|_2 \|H^T\|_2} \right|
\end{equation*}

where
\( H H^T \) computes the pairwise dot products between layer hidden states. The denominator normalizes the values using the \( L_2 \)-norm, defined as:

\begin{equation*}
\|H_i\|_2 = \sqrt{\sum_{k=1}^{d} (H_i^{(k)})^2}
\end{equation*}
ensuring that values lie within \( [-1,1] \). The absolute value guarantees that all elements are within the range \( [0,1] \). Then, we compute the average of \(S\) across four tasks.

We also explore the variations in visual representation spaces across four tasks. As illustrated in Figure~\ref{fig:cosine_similarity_clip_4}, the partitioning of visual representations is minimally influenced by the nature of the tasks. In other words, the shallow, middle, and deep representation spaces exhibit remarkable stability, maintaining consistent structures across various tasks.

\subsection{Replacing the Penultimate Layer}
As shown in Table~\ref{tab:notraining_llava}, when the penultimate-layer visual representations are replaced by those from other layers of the visual encoder without additional training, layers 20 to 24 (belonging to the deep representation space) do not suffer catastrophic performance degradation. Moreover, among the two layers closest to the penultimate one, layer 22 exhibits more stable performance than the final layer.


\begin{table}[htbp]
\centering
\resizebox{0.5\textwidth}{!}{ 
\begin{tabular}{@{}c cccccc@{}}
\toprule
\rowcolor[HTML]{EFEFEF} 
\textbf{Layer} & \textbf{MME-P} & \textbf{MME-C} & \textbf{OCRB} & \textbf{TextVQA} & \textbf{RefCOCO} \\
\midrule
24 & 1153.5 & 306.0 & 266 & 41.16 & 47.56 \\
23 & 1509.9 & 365.3 & 314 & 46.10 & 49.04 \\
22 & 1451.1 & 366.7 & 304 & 44.76 & 47.46 \\
21 & 1368.8 & 293.2 & 287 & 41.59 & 40.47 \\
20 & 1259.2 & 265.7 & 271 & 39.11 & 44.31 \\
19 & 1183.3 & 267.1 & 240 & 36.76 & 42.83 \\
18 & 1083.7 & 237.8 & 205 & 32.18 & 36.04 \\
17 & 993.6  & 255.7 & 156 & 27.92 & 31.07 \\
16 & 901.0  & 256.7 & 116 & 23.37 & 19.96 \\
15 & 790.0  & 253.9 & 94  & 17.96 & 14.85 \\
\bottomrule
\end{tabular}
}
\caption{\small Performance metrics across different layers on various benchmarks for non-training methods are presented. Specifically, MME-P denotes MME Perception, MME-C corresponds to MME Cognition, and OCRB represents OCRBench. The performance on RefCOCO is evaluated using Intersection over Union (IOU) as the metric.}
\label{tab:notraining_llava}
\end{table}

\begin{table*}[ht]

\label{tab:detail_ft}
\centering \footnotesize
  \renewcommand\tabcolsep{15.6pt}
\begin{tabular}{l l}
\toprule
Data & Size \\
\midrule
\textbf{LLaVA}~\cite{liu2024visual} & 158K \\
\midrule
+ ShareGPT~\cite{sharegpt} & 40K \\
+ VQAv2~\cite{goyal2017vqav2} & 83K \\
+ GQA~\cite{hudson2019gqa} & 72K \\
+ OKVQA~\cite{okvqa} & 9K \\
+ OCRVQA~\cite{mishra2019ocrvqa} & 80K \\
+ A-OKVQA~\cite{schwenk2022okvqa} & 66K  \\
+ TextCaps~\cite{sidorov2020textcaps} & 22K  \\
+ RefCOCO \cite{kazemzadeh2014referitgame,mao2016generation} & 48K \\
+ VG~\cite{krishna2017visual} & 86K  \\
\midrule
\textbf{LLaVA-1.5}~\cite{liu2023improvedllava} & 665K  \\
\midrule
+ AI2D~\cite{kembhavi2016diagram} & 16K \\
+ DocVQA~\cite{mathew2021docvqa} & 15K \\
+ DVQA~\cite{kafle2018dvqa} & 13K \\
\midrule
\textbf{Cambrian-737k}~\cite{tong2024cambrian} & 737K  \\
\midrule
+ CLEVR~\cite{johnson2017clevr} & 215k \\
+ TallyQA~\cite{acharya2019tallyqa} & 77K \\
\textbf{Customized-1M} & 1M \\

\bottomrule
\end{tabular}
\caption{
\small The mixture detail of fine-tuning dataset for LLaVA-1.5 665K, Cambrian-1 737K and customized 1M.
}
\label{appendix:data_composition}
\end{table*}

\section{Experiment Details}
\label{appendix:exp_details}

\subsection{Composition of Three Scale Datasets}
As shown in Table~\ref{appendix:data_composition}, the following datasets are incorporated to enhance the model’s capabilities across multiple multimodal tasks, with all datasets accessed and utilized strictly under their licenses.

\begin{itemize}
    \item \textbf{AI2D (Allen Institute for AI Diagram Dataset) \cite{kembhavi2016diagram}}  
    AI2D is designed for visual reasoning and diagram understanding, featuring annotated diagrams with textual descriptions and Q\&A pairs. It is particularly useful for multimodal reasoning and visual question answering (VQA) tasks.
    \item \textbf{DocVQA (Document Visual Question Answering) \cite{mathew2021docvqa}}  
    DocVQA focuses on visual question answering over document images, where questions pertain to scanned documents, OCR-recognized text, and textual reasoning. This dataset is valuable for document comprehension, text recognition, and multimodal reasoning.
    \item \textbf{DVQA (Diagrammatic Visual Question Answering) ~\cite{kafle2018dvqa}}  
    DVQA is designed for visual question answering over diagrams and charts, covering questions related to bar charts, pie charts, and scientific illustrations. It evaluates the model's ability to read structured visual information and perform reasoning based on graphical representations.
    \item \textbf{CLEVR (Compositional Language and Elementary Visual Reasoning) ~\cite{johnson2017clevr}}  
    CLEVR is a synthetic dataset for visual reasoning, containing 3D-rendered scenes with structured questions that require reasoning based on attributes, object relationships, and compositional logic. It is widely used to assess a model’s capability in compositional and multi-step reasoning.
    \item \textbf{TallyQA ~\cite{acharya2019tallyqa}}  
    TallyQA is a dataset specifically designed for object counting tasks, where questions require the model to accurately count objects in an image. It evaluates the model's ability to attend to relevant objects, integrate global and local information, and perform numerical reasoning.
    
\end{itemize}
\subsection{Evaluation Metrics}
We provide a comprehensive explanation of the evaluation methods, categorizing them into three distinct types based on the evaluation metrics:
\begin{itemize}
    \item For benchmarks such as MME-Perception, MME-Cognition, OCRBench, and MMVet, we adopt the common approach of directly using the dataset-defined scores. We follow this established approach to maintain consistency and comparability in evaluations.
    \item Using Accuracy directly as the evaluation metric. This applies to benchmarks such as MMBench, SEEDBench, GQA, TextVQA, CVBench, RealworldQA, and POPE.
    \item In evaluating the RefCOCO dataset, we use CIDEr (Consensus-based Image Description Evaluation) as the primary evaluation metric.
\end{itemize}


To facilitate evaluation, we use lmms-eval as our primary evaluation tool. Evaluations must be conducted on the official platform for the MMVet dataset by uploading the necessary data. Regarding the CVBench 3D tasks, where models generally exhibit weaker instruction-following performance, we employ the DeepSeek API as the judge. This tool provides results consistent with GPT-4o but is significantly more cost-effective.

\subsection{The impact of LLMs}
\label{appendix:llm_size}
Additional experiments are conducted on different sizes of large language models to investigate their impact on visual information processing. We validate our conclusions on LLMs of 2.7b and 7b sizes. Due to computational resource constraints, we selected representative layers from the three representation spaces to conduct experiments on subtasks of MME and SEEDBench.
As shown in Table~\ref{tab:mme-subtasks-2.7b-7b}, the penultimate layer does not consistently achieve the best performance on MME. The commonly used penultimate layer achieves optimal performance on 6 out of 14 subtasks, while other layers, such as Layers 3, 18, and 24, demonstrate superior performance on the remaining subtasks. This observation aligns with prior findings, suggesting that middle layers can exhibit superior performance over deeper layers on certain CV-centric tasks. Notably, layer 18 outperforms the penultimate layer in tasks such as Count, Position, and Existence.

\begin{table*}[ht]
\centering
\renewcommand{\arraystretch}{1.2} 
\setlength{\tabcolsep}{4pt}    
\resizebox{0.6\textwidth}{!}{
\begin{tabular}{lcccc|cccc}
\toprule
\textbf{Model Size} & \multicolumn{4}{c|}{\textbf{2.7b}} & \multicolumn{4}{c}{\textbf{7b}} \\ 
\cmidrule(r){2-5} \cmidrule(r){6-9}
\textbf{Layers}
                   & \textbf{3} & \textbf{18} & \textbf{23} & \textbf{24} & \textbf{3} & \textbf{18} & \textbf{23} & \textbf{24} \\ 
\midrule
\textbf{Code Reasoning}         & \textbf{52.50} & 47.50 & 47.50 & 40.00 & \textbf{50.00} & 40.00 & 42.50 & 45.00 \\ 
\textbf{Artwork}                & 53.00 & 65.00 & \textbf{65.75} & 64.50 & 50.00 & 69.25 & \textbf{71.00} & 70.75 \\ 
\textbf{Celebrity}              & 46.76 & 49.12 & \textbf{64.12} & 58.82 & 51.76 & 59.41 & \textbf{74.71} & 74.41 \\ 
\textbf{Numerical Calculation}  & \textbf{50.00} & \textbf{50.00} & 42.50 & 25.00 & \textbf{47.50} & 45.00 & 37.50 & 37.50 \\ 
\textbf{Text Translation}       & 50.00 & 50.00 & 50.00 & 50.00 & 65.00 & 65.00 & 47.50 & \textbf{67.50} \\ 
\textbf{Count}                  & 50.00 & \textbf{65.00} & 61.67 & 58.33 & 56.67 & \textbf{85.00} & \textbf{85.00} & 80.00 \\ 
\textbf{Color}                  & 53.33 & 83.33 & 86.67 & \textbf{88.33} & 78.33 & \textbf{91.67} & \textbf{91.67} & \textbf{91.67} \\ 
\textbf{Commonsense Reasoning}  & 52.86 & 60.71 & \textbf{64.29} & 62.14 & 57.86 & 69.29 & \textbf{73.57} & 72.86 \\ 
\textbf{Position}               & 48.33 & \textbf{71.67} & \textbf{71.67} & \textbf{71.67} & 61.67 & 71.67 & \textbf{75.00} & 80.00 \\ 
\textbf{OCR}                    & 50.00 & 67.50 & \textbf{72.50} & 65.00 & 55.00 & \textbf{77.50} & 75.00 & 70.00 \\ 
\textbf{Landmark}               & 59.50 & 75.25 & \textbf{80.25} & 76.75 & 66.25 & 78.00 & \textbf{86.00} & 84.50 \\ 
\textbf{Scene}                  & 73.75 & 87.75 & 87.75 & \textbf{89.00} & 80.00 & 85.50 & 85.50 & \textbf{87.50} \\ 
\textbf{Existence}              & 83.33 & \textbf{98.33} & 96.67 & 95.00 & 81.67 & \textbf{96.67} & \textbf{96.67} & 95.00 \\ 
\textbf{Posters}                & 37.41 & 59.18 & \textbf{65.65} & 64.29 & 51.02 & 74.83 & 81.63 & \textbf{83.33} \\ 
\bottomrule
\end{tabular}
}
\caption{\small Performance of LLaVA architectures with 2.7B and 7B LLMs on MME subtasks, evaluated across four layers from three representative spaces.}
\label{tab:mme-subtasks-2.7b-7b}
\end{table*}
\begin{table*}[ht]
\centering
\renewcommand{\arraystretch}{1.3} 
\setlength{\tabcolsep}{10pt}  
\resizebox{0.8\textwidth}{!}{
\begin{tabular}{lcccccccc}
\toprule
\textbf{Model Size} & \multicolumn{4}{c}{\textbf{2.7b}} & \multicolumn{4}{c}{\textbf{7b}} \\ 
\cmidrule(r){2-5} \cmidrule(r){6-9}
\textbf{Layers} & \textbf{3} & \textbf{18} & \textbf{23} & \textbf{24} & \textbf{3} & \textbf{18} & \textbf{23} & \textbf{24} \\ 
\midrule
\textbf{Scene Understanding}   & 36.04  & 68.84  & 68.87  & \textbf{69.79} & 50.70  & 73.50  & \textbf{73.91} & 73.59  \\ 
\textbf{Instance Identity}     & 32.33  & 62.26  & \textbf{62.92}  & 62.59  & 41.40  & 67.78  & 70.29  & \textbf{70.56}  \\ 
\textbf{Instance Attribute}    & 40.35  & \textbf{62.19}  & 59.07  & 60.46  & 50.48  & 69.09  & \textbf{68.70}  & 68.21  \\ 
\textbf{Instance Location}     & 37.53  & 52.25  & 49.69  & \textbf{53.78}  & 43.46  & 61.04  & \textbf{61.45}  & 59.71  \\ 
\textbf{Instance Counting}     & 25.70  & 43.07  & \textbf{47.57}  & 45.20  & 33.02  & 56.89  & 57.13  & \textbf{57.29}  \\ 
\textbf{Spatial Relation}      & 33.03  & 42.47  & 40.64  & \textbf{43.99}  & 41.55  & \textbf{52.05}  & 49.62  & 51.45  \\ 
\textbf{Instance Interaction}  & 34.02  & \textbf{64.95}  & 54.64  & 64.95  & 48.45  & 63.92  & 67.01  & \textbf{71.13}  \\ 
\textbf{Visual Reasoning}      & 35.05  & 67.07  & \textbf{72.51}  & 72.21  & 53.47  & 75.23  & \textbf{78.85}  & 77.04  \\ 
\textbf{Text Recognition}      & \textbf{44.71}  & 21.18  & 21.18  & 24.71  & 38.82  & 34.12  & \textbf{47.06}  & 43.53  \\ 
\bottomrule
\end{tabular}
}
\caption{\small Performance of LLaVA architectures with 2.7B and 7B LLMs on SEEDBench subtasks, evaluated across four layers from three representative spaces.}
\label{tab:seedbench-2.7-7b}
\end{table*}
As illustrated in Table~\ref{tab:seedbench-2.7-7b}, while performance varies slightly across the subtasks of SEEDBench, the penultimate layer achieves the best performance on only 3 out of 9 subtasks. These results provide strong empirical evidence that shallow and middle layers can outperform deeper layers on specific subtasks. 

\subsection{The impact of Data Scale}
\label{appendix:data_scale}
The conclusion remains valid across different data scales. Under the 737k data scale, half of the subtasks in the MME dataset achieve optimal performance using the penultimate layer. However, for tasks like Count, Position, and Existence, the middle visual representation layer (Layer 18) demonstrates either superior or comparable performance. Similarly, results under the 1M data scale also show that half of the optimal performances are achieved on layers other than the penultimate one. 

The results for SEEDBench subtasks, as presented in Table~\ref{tab:seedbench-737k-1M}, further support this observation. At the 737k data scale, Layer 18 from the middle representation space achieves the best performance on 5 out of 9 subtasks, while the penultimate layer excels in only 3 subtasks. Likewise, under the 1M data scale, half of the subtasks continue to achieve their best performance on layers other than the penultimate one. These findings consistently demonstrate across varying data scales that shallow and middle layers have the potential to outperform deep layers in certain scenarios.

\subsection{Layer Selections and Feature Fusion}
\begin{table*}[t]
    \centering
    \renewcommand\tabcolsep{2.4pt}

    \resizebox{\textwidth}{!}{ 
    \begin{tabular}{lllllllllllllc}
        \toprule
        \multirow{2}{*}{\textbf{Models}} & \multicolumn{5}{c}{\textbf{General}} & \multicolumn{2}{c}{\textbf{OCR}} & \multicolumn{4}{c}{\textbf{Vision-Centric}} & \multicolumn{1}{c}{\textbf{Hallu}} \\
        \cmidrule(lr){2-6} \cmidrule(lr){7-8} \cmidrule(lr){9-12} \cmidrule(lr){13-13}
        & MME$^P$ & MME$^C$ & MMB & SEEDB & GQA & TextVQA & OCRB & CVB & CVB$^{2D}$ & CVB$^{3D}$ & RWQA & POPE &  \textbf{Win} \\
        \midrule
        \rowcolor{gray!15}{Baseline(${23}$)}   & 1142.8  & 245.0  & 35.31  & 52.84  & 52.84  & 33.73  & 233  & 44.26  & 38.02  & 50.50  & 45.36  & 84.00 &\textcolor{PineGreen}{9/10} \\
        {+ \textit{18}}     & 1148.5$^{\color{red}\text{\scriptsize 5.7$\uparrow$}}$  & 228.9$^{\color{teal}\text{\scriptsize 16.1$\downarrow$}}$  & 46.91$^{\color{red}\text{\scriptsize 11.6$\uparrow$}}$  & 57.01$^{\color{red}\text{\scriptsize 4.2$\uparrow$}}$  & 56.80$^{\color{red}\text{\scriptsize 4$\uparrow$}}$  & 37.66$^{\color{red}\text{\scriptsize 3.9$\uparrow$}}$  & 273$^{\color{red}\text{\scriptsize 40$\uparrow$}}$  & 44.73$^{\color{red}\text{\scriptsize 0.5$\uparrow$}}$  & 35.79$^{\color{teal}\text{\scriptsize 2.2$\downarrow$}}$  & 53.67$^{\color{red}\text{\scriptsize 3.2$\uparrow$}}$  & 45.49$^{\color{red}\text{\scriptsize 0.1$\uparrow$}}$  & 84.51$^{\color{red}\text{\scriptsize 0.5$\uparrow$}}$  &\textcolor{PineGreen}{8/10}\\
        {+ \textit{1+18}}  & 1155.4$^{\color{red}\text{\scriptsize 12.6$\uparrow$}}$  & 246.8$^{\color{red}\text{\scriptsize 1.8$\uparrow$}}$  & 48.54$^{\color{red}\text{\scriptsize 13.2$\uparrow$}}$  & 56.75$^{\color{red}\text{\scriptsize 3.9$\uparrow$}}$  & 56.68$^{\color{red}\text{\scriptsize 3.8$\uparrow$}}$  & 36.53$^{\color{red}\text{\scriptsize 2.8$\uparrow$}}$  & 236$^{\color{red}\text{\scriptsize 3$\uparrow$}}$  & 45.65$^{\color{red}\text{\scriptsize 1.4$\uparrow$}}$  & 36.21$^{\color{teal}\text{\scriptsize 1.8$\downarrow$}}$  & 55.08$^{\color{red}\text{\scriptsize 4.6$\uparrow$}}$  & 46.93$^{\color{red}\text{\scriptsize 1.6$\uparrow$}}$  & 84.56$^{\color{red}\text{\scriptsize 0.6$\uparrow$}}$  &\textcolor{PineGreen}{7/10}\\
        {+ \textit{17+18}} & 1182.5$^{\color{red}\text{\scriptsize 39.7$\uparrow$}}$  & 220.7$^{\color{teal}\text{\scriptsize 24.3$\downarrow$}}$  & 48.80$^{\color{red}\text{\scriptsize 13.5$\uparrow$}}$  & 56.68$^{\color{red}\text{\scriptsize 3.8$\uparrow$}}$  & 56.48$^{\color{red}\text{\scriptsize 3.6$\uparrow$}}$  & 38.29$^{\color{red}\text{\scriptsize 4.6$\uparrow$}}$  & 263$^{\color{red}\text{\scriptsize 30$\uparrow$}}$  & 45.38$^{\color{red}\text{\scriptsize 1.1$\uparrow$}}$  & 36.25$^{\color{teal}\text{\scriptsize 1.8$\downarrow$}}$  & 54.50$^{\color{red}\text{\scriptsize 4$\uparrow$}}$  & 44.71$^{\color{teal}\text{\scriptsize 0.7$\downarrow$}}$  & 85.50$^{\color{red}\text{\scriptsize 1.5$\uparrow$}}$ &\textcolor{PineGreen}{6/10} \\
        \midrule
        {DC-STI}  & 1142.4$^{\color{teal}\text{\scriptsize 0.4$\downarrow$}}$  & 218.9$^{\color{teal}\text{\scriptsize 26.1$\downarrow$}}$  & 48.02$^{\color{red}\text{\scriptsize 12.7$\uparrow$}}$  & 57.23$^{\color{red}\text{\scriptsize 4.4$\uparrow$}}$  & 56.86$^{\color{red}\text{\scriptsize 4.0$\uparrow$}}$  & 36.42$^{\color{red}\text{\scriptsize 2.7$\uparrow$}}$  & 226$^{\color{teal}\text{\scriptsize 7$\downarrow$}}$  & 43.83$^{\color{teal}\text{\scriptsize 0.4$\downarrow$}}$  & 35.10$^{\color{teal}\text{\scriptsize 2.9$\downarrow$}}$  & 52.58$^{\color{red}\text{\scriptsize 2.1$\uparrow$}}$  & 44.44$^{\color{teal}\text{\scriptsize 0.9$\downarrow$}}$  & 86.38$^{\color{red}\text{\scriptsize 2.4$\uparrow$}}$  &\textcolor{PineGreen}{8/10}\\
        \({\text{DC-SCI}}^{*}\)  & 1166.5$^{\color{red}\text{\scriptsize 23.7$\uparrow$}}$  & 241.8$^{\color{teal}\text{\scriptsize 3.2$\downarrow$}}$  & 48.71$^{\color{red}\text{\scriptsize 13.4$\uparrow$}}$  & 57.26$^{\color{red}\text{\scriptsize 4.4$\uparrow$}}$  & 56.61$^{\color{red}\text{\scriptsize 3.8$\uparrow$}}$  & 36.70$^{\color{red}\text{\scriptsize 3.0$\uparrow$}}$  & 241$^{\color{red}\text{\scriptsize 8$\uparrow$}}$  & 43.19$^{\color{teal}\text{\scriptsize 1.1$\downarrow$}}$  & 34.96$^{\color{teal}\text{\scriptsize 3.1$\downarrow$}}$  & 51.42$^{\color{red}\text{\scriptsize 0.9$\uparrow$}}$  & 44.44$^{\color{teal}\text{\scriptsize 0.9$\downarrow$}}$  & 84.45$^{\color{red}\text{\scriptsize 0.5$\uparrow$}}$ &\textcolor{PineGreen}{7/10} \\
        {DC-DCI}  & 1145.0$^{\color{red}\text{\scriptsize 2.2$\uparrow$}}$  & 253.2$^{\color{red}\text{\scriptsize 8.2$\uparrow$}}$  & 47.85$^{\color{red}\text{\scriptsize 12.5$\uparrow$}}$  & 57.16$^{\color{red}\text{\scriptsize 4.3$\uparrow$}}$  & 56.92$^{\color{red}\text{\scriptsize 4.1$\uparrow$}}$  & 37.54$^{\color{red}\text{\scriptsize 3.8$\uparrow$}}$  & 257$^{\color{red}\text{\scriptsize 24$\uparrow$}}$  & 45.60$^{\color{red}\text{\scriptsize 1.3$\uparrow$}}$  & 35.83$^{\color{teal}\text{\scriptsize 2.2$\downarrow$}}$  & 54.92$^{\color{red}\text{\scriptsize 4.4$\uparrow$}}$  & 45.10$^{\color{teal}\text{\scriptsize 0.3$\downarrow$}}$  & 84.95$^{\color{red}\text{\scriptsize 1.0$\uparrow$}}$  &\textcolor{PineGreen}{7/10} \\
        {MMFuser}  & 1149.5$^{\color{red}\text{\scriptsize 6.7$\uparrow$}}$  & 238.9$^{\color{teal}\text{\scriptsize 6.1$\downarrow$}}$  & 49.65$^{\color{red}\text{\scriptsize 14.3$\uparrow$}}$  & 56.21$^{\color{red}\text{\scriptsize 3.4$\uparrow$}}$  & 56.59$^{\color{red}\text{\scriptsize 3.8$\uparrow$}}$  & 35.43$^{\color{red}\text{\scriptsize 1.7$\uparrow$}}$  & 245$^{\color{red}\text{\scriptsize 12$\uparrow$}}$  & 45.70$^{\color{red}\text{\scriptsize 1.4$\uparrow$}}$  & 36.89$^{\color{teal}\text{\scriptsize 1.1$\downarrow$}}$  & 54.50$^{\color{red}\text{\scriptsize 4.0$\uparrow$}}$  & 44.83$^{\color{teal}\text{\scriptsize 0.5$\downarrow$}}$  & 84.53$^{\color{red}\text{\scriptsize 0.5$\uparrow$}}$ &\textcolor{PineGreen}{8/10} \\
        \rowcolor{gray!15}\({\text{Ours}}^{*}\)  & 1157.2$^{\color{red}\text{\scriptsize 14.4$\uparrow$}}$  & 236.1$^{\color{teal}\text{\scriptsize 8.9$\downarrow$}}$  & 49.22$^{\color{red}\text{\scriptsize 13.9$\uparrow$}}$  & 57.23$^{\color{red}\text{\scriptsize 4.4$\uparrow$}}$  & 57.35$^{\color{red}\text{\scriptsize 4.5$\uparrow$}}$  & 37.70$^{\color{red}\text{\scriptsize 4.0$\uparrow$}}$  & 265$^{\color{red}\text{\scriptsize 32$\uparrow$}}$  & 44.56$^{\color{red}\text{\scriptsize 0.3$\uparrow$}}$  & 36.53$^{\color{teal}\text{\scriptsize 1.5$\downarrow$}}$  & 52.58$^{\color{red}\text{\scriptsize 2.1$\uparrow$}}$  & 45.75$^{\color{red}\text{\scriptsize 0.4$\uparrow$}}$  & 84.82$^{\color{red}\text{\scriptsize 0.8$\uparrow$}}$ & - \\
        \bottomrule
    \end{tabular}
    }
        \caption{\small Study on different layer fusion strategies. The results reveal that nearly all fusion methods significantly outperform the baseline, with performance variations depending on the combination of different layers.  (*) 'DC-SCI' is the same as \(\mathcal{L}_4\) and  'Ours' represents \(\mathcal{L}_5\). 'Win' denotes the proportion of datasets where our method achieves superior performance.}
    \label{tab:exp3-2_table_compare_appendix}
\end{table*}

In the shallow layer, we select layers 1 and 3 as representatives. Layer 1, being the most chaotic, primarily captures early-stage visual features, while layer 3 is still in a chaotic state but performs relatively well. In the middle layer, we choose layers 18 and 17, as they achieve the first and second-best performance within this representation space. For the deep layer, we select layer 23, as it demonstrates the highest overall performance. The baseline configuration considers only layer 23 as the visual representation.

Full version of comparison study as shown in Figure~\ref{tab:exp3-2_table_compare_appendix}. We evaluate four state-of-the-art fusion methods, including three from DenseConnector (DC) ~\cite{yao2024dense} and one from MMFuser ~\cite{cao2024mmfuser}. As shown in Table~\ref{tab:exp3-2_table_compare}, our method outperforms STI, SCI, and DCI on 8, 7, and 7 benchmarks, respectively. Compared to MMFuser ~\cite{cao2024mmfuser}, our approach demonstrates superior performance on 8 benchmarks. These results highlight the significant potential of visual feature fusion strategies in enhancing MLLMs and offering guidance for developing future fusion strategies.
\begin{table*}[ht]
\centering
\renewcommand{\arraystretch}{1.2} 
\setlength{\tabcolsep}{4pt}     
\resizebox{0.6\textwidth}{!}{
\begin{tabular}{lcccc|cccc}
\toprule
\textbf{Data Scale} & \multicolumn{4}{c|}{\textbf{737k}} & \multicolumn{4}{c}{\textbf{1M}} \\ 
\cmidrule(r){2-5} \cmidrule(r){6-9}
\textbf{Layers}
                   & \textbf{3} & \textbf{18} & \textbf{23} & \textbf{24} & \textbf{3} & \textbf{18} & \textbf{23} & \textbf{24} \\ 
\midrule
\textbf{Code Reasoning}         & \textbf{50.00} & 47.50 & 45.00 & 42.50 & \textbf{47.50} & \textbf{47.50} & 45.00 & \textbf{47.50} \\ 
\textbf{Artwork}                & 51.00 & 59.25 & \textbf{64.50} & 61.75 & 53.75 & 65.00 & \textbf{68.00} & 66.50 \\ 
\textbf{Celebrity}              & 48.82 & 55.88 & \textbf{64.12} & 62.65 & 52.35 & 62.06 & \textbf{68.53} & 65.59 \\ 
\textbf{Numerical Calculation}  & \textbf{47.50} & 37.50 & 35.00 & \textbf{47.50} & \textbf{50.00} & 30.00 & 45.00 & 47.50 \\ 
\textbf{Text Translation}       & 50.00 & 50.00 & 50.00 & 50.00 & 50.00 & 50.00 & 50.00 & 50.00 \\ 
\textbf{Count}                  & \textbf{55.00} & \textbf{55.00} & \textbf{55.00} & 50.00 & 58.33 & 58.33 & \textbf{60.00} & \textbf{60.00} \\ 
\textbf{Color}                  & 60.00 & 75.00 & \textbf{76.67} & 75.00 & 61.67 & 78.33 & \textbf{80.00} & \textbf{80.00} \\ 
\textbf{Commonsense Reasoning}  & 54.29 & 56.43 & 60.71 & \textbf{61.43} & 52.14 & 57.14 & \textbf{62.14} & 58.57 \\ 
\textbf{Position}               & 51.67 & \textbf{70.00} & 63.33 & \textbf{70.00} & 50.00 & 70.00 & \textbf{73.33} & 70.00 \\ 
\textbf{OCR}                    & 50.00 & 55.00 & \textbf{57.50} & 52.50 & \textbf{55.00} & 52.50 & \textbf{55.00} & \textbf{55.00} \\ 
\textbf{Landmark}               & 63.25 & 71.00 & \textbf{77.50} & 74.25 & 61.75 & 72.50 & \textbf{77.50} & 76.00 \\ 
\textbf{Scene}                  & 69.25 & 83.50 & \textbf{85.00} & 84.25 & 74.00 & 84.50 & 84.00 & \textbf{85.75} \\ 
\textbf{Existence}              & 85.00 & \textbf{96.67} & \textbf{96.67}& \textbf{96.67} & 80.00 & \textbf{96.67} & \textbf{96.67} & 95.00 \\ 
\textbf{Posters}                & 38.44 & 50.34 & 55.44 & \textbf{56.80} & 37.41 & 52.72 & \textbf{55.44} & 54.76 \\ 
\bottomrule
\end{tabular}
}
\caption{\small Experimental results on MME subtasks under data scales of 737k and 1M, with the model settings consistent with those described in the main text.}
\label{tab:MME-737k-1M}
\end{table*}

\begin{table*}[ht]
\centering
\renewcommand{\arraystretch}{1.2} 
\setlength{\tabcolsep}{4pt}   
\resizebox{0.6\textwidth}{!}{ 
\begin{tabular}{lcccccccc}
\toprule
\textbf{Data Scale} & \multicolumn{4}{c}{\textbf{737k}} & \multicolumn{4}{c}{\textbf{1M}} \\ \cmidrule(lr){2-5} \cmidrule(lr){6-9}
\textbf{Layers}
 & \textbf{3} & \textbf{18} & \textbf{23} & \textbf{24} & \textbf{3} & \textbf{18} & \textbf{23} & \textbf{24} \\ \midrule
\textbf{Scene Understanding}   & 26.85 & 67.35 & \textbf{68.24} & 67.57 & 44.02 & 65.71 & \textbf{68.97}& 67.57 \\
\textbf{Instance Identity}     & 23.65 & \textbf{56.36} & 55.60 & 55.87 & 35.06 & 57.89 & \textbf{61.82} & 58.77 \\
\textbf{Instance Attribute}    & 26.03 & \textbf{59.93} & 58.36 & 57.69 & 44.27 & \textbf{61.93} & 61.45 & 60.59 \\
\textbf{Instance Location}     & 26.69 & \textbf{50.20} & 47.75 & 48.26 & 39.37 & 48.67 & \textbf{49.69} & 49.08 \\
\textbf{Instance Counting}     & 27.34 & \textbf{41.93} & 40.13 & 41.19 & 32.04 & 42.38 & 44.91 & \textbf{45.97} \\
\textbf{Spatial Relation}     & 29.68 & \textbf{41.86} & 38.81 & 38.96 & 37.90 & \textbf{42.92} & 39.57 & 39.57 \\
\textbf{Instance Interaction}  & 21.65 & 46.39 & \textbf{51.55} & 49.48 & 38.14 & 50.52 & 51.55 & \textbf{52.58} \\
\textbf{Visual Reasoning}      & 27.79 & 62.54 & 67.98 & \textbf{69.18} & 43.20 & 60.42 & \textbf{68.88} & 67.07 \\
\textbf{Text Recognition}      & 21.18 & 34.12 & \textbf{48.24} & 43.53 & 41.18 & 17.65 & \textbf{55.29} & 42.35 \\ \bottomrule
\end{tabular}}
\caption{\small Experimental results on SEEDBench subtasks under data scales of 737k and 1M, with the model settings consistent with those described in the main text.}
\label{tab:seedbench-737k-1M}
\end{table*}

\begin{figure*}[hb]
    \centering
    \includegraphics[width=1\linewidth]{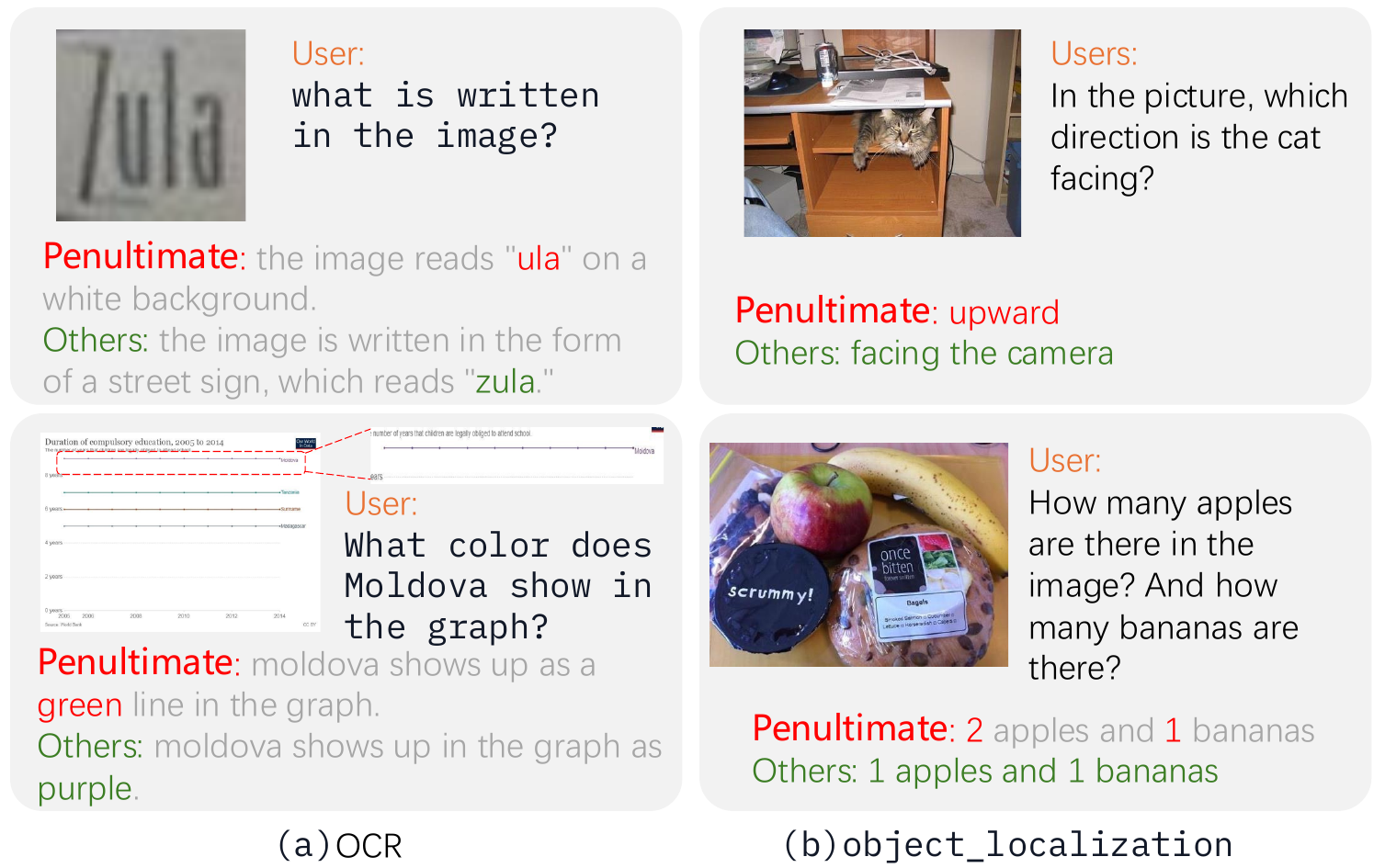}
    \caption{\small Case study illustrating four examples where the penultimate layer provides incorrect answers, but these errors can be resolved by using shallow and middle layers. In all four cases, Layer 18 of CLIP-ViT as the visual representation could successfully provide the correct answers.}
    \label{fig:samplesl}
\end{figure*}

\begin{figure*}
    \centering
    \includegraphics[width=1\linewidth]{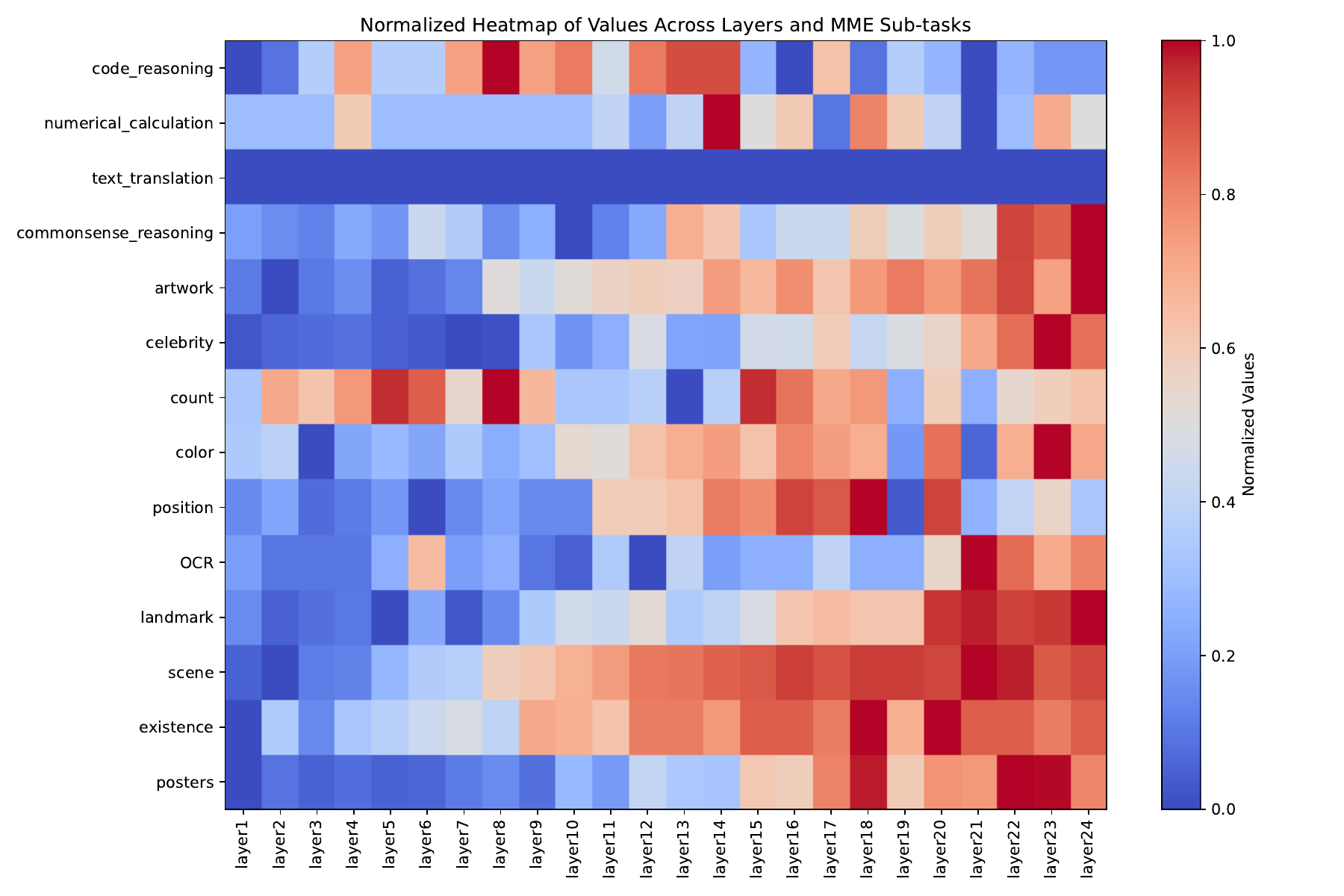}
    \caption{\small Heatmap showing the relative performance of CLIP-ViT layers 1-24 on MME subtasks. Consistent with the experimental settings in the main text, the results demonstrate that while deep layers generally achieve the best performance, shallow and middle layers can surpass deep layers on specific tasks, providing further support for our findings.}
    \label{fig:mme_layer_wise_heatmap}
\end{figure*}

\section{Detailed Discussion}
\label{appendix:C}
\subsection{Alignment Guideline}
\textit{What defines a good visual representation in multimodal models?} Firstly, an ideal visual representation must simultaneously provide rich visual information and effectively align with textual modal. An MLLM can only correctly answer queries when the information corresponding to the given instruction is explicitly embedded within the visual representation. However, when the visual representation fails to deliver the necessary information, the model becomes prone to hallucination problems. Secondly, alignment with the textual modality is essential for a large language model to understand and process information from a different modality. This alignment ensures that the rich visual content is effectively leveraged. In a word, the CLIP series models currently offer the best trade-off between these two dimensions.

\subsection{Comparison with Prior Work}
Although \cite{liu2023llava,yao2024dense,cao2024mmfuser,li2026instruction} have briefly considered features from different layers, none have conducted a systematic or in-depth analysis of layer selection or the functional roles of visual layers in MLLMs.
\begin{table*}[hb]
\centering
\begin{tabular}{lccccc}
\hline
\textbf{Model} & \textbf{MMB} & \textbf{SEEDB} & \textbf{GQA} & \textbf{OCRB} & \textbf{RefCOCO} \\
\hline
Baseline (Vicuna-7B) & 65.1 & 65.2 & 63.2 & 320 & 52.3 \\
DenseConnector~\cite{yao2024dense} & 63.3 & 66.6 & 62.9 & 323 & - \\
MMFuser~\cite{cao2024mmfuser} & 67.5 & 60.8 & 62.8 & - & - \\
Instruction-Guided Fusion~\cite{li2026instruction} & 66.9 & 68.3 & 63.1 & - & - \\
Ours & 67.4 & 67.7 & 63.5 & 325 & 54.0 \\
\hline
\end{tabular}
\caption{\small Performance comparison on 7B-scale LLMs. Dense2Connector and MMFuser are retrained under the same settings, while results of Instruction-Guided Fusion are copied from \cite{li2026instruction}. The results highlight the importance of visual layer selection.}
\label{tab:benchmark_results}
\end{table*}

\begin{itemize}
    \item \cite{liu2023llava} presents a limited comparison between the penultimate and final layers on ScienceQA, without broader analysis. In contrast, we generalize this observation across over 10 benchmarks and various model and data scales, and offer an interpretation in Section~\ref{sec:exp1-2} grounded in CLIP’s contrastive learning objective.
    \item \cite{yao2024dense} uses attention maps to suggest that different layers attend to different regions of interest. However, such analysis is inherently qualitative and subjective, lacking a principled explanation of why these differences arise or how they relate to downstream tasks.
    \item \cite{cao2024mmfuser} proposes a shallow-to-deep fusion strategy but does not explain the underlying rationale for selecting specific layers, and its choices are based on empirical heuristics (e.g., 1/3, 2/3, 3/3), with no analysis of representations itself.
    \item While \cite{li2026instruction} focuses on fusion design, it lacks consideration of the role of visual layers in MLLMs. Our work provides novel insights, such as the unexpected correlation between inter-layer similarity and downstream performance. Moreover, they analyze only 1/5 of ViT layers, without considering data scale or model size. 
\end{itemize}

In contrast, our work is the first to systematically group ViT layers based on representation similarity, and to establish clear correlations between layer types and multimodal task performance across 60+ tasks and multiple model sizes (1.4B–7B) and data scales (665k-1M).

\section{Ethics Statement}
\label{Appendix:D}
This work made limited use of ChatGPT for language polishing and code debugging support. All scientific contributions, analyses, and conclusions are those of the authors.

\begin{table*}[t]
\centering

\renewcommand{\arraystretch}{1.3}
\setlength{\tabcolsep}{3pt} 
\begin{tabular}{>{\centering\arraybackslash}m{1cm}
                >{\centering\arraybackslash}m{3.6cm}
                >{\centering\arraybackslash}m{3.6cm}
                >{\centering\arraybackslash}m{3.1cm}
                >{\centering\arraybackslash}m{3.1cm}}
\toprule
\textbf{Index} & \textbf{MME} & \textbf{MMVet} & \textbf{MMBench} & \textbf{SEEDBench} \\
\midrule
1  & Code Reasoning        & Rec\_OCR\_Spat\_Math & Attribute Comparison      & Instance Attribute \\
2  & Count                 & Rec\_Spat            & Attribute Recognition     & Instance Location \\
3  & Numerical Calculation & Rec\_OCR\_Know\_Gen  & Physical Property         & Instance Interaction \\
4  & Position              & OCR\_Spat            & Spatial Relation          & Text Recognition \\
5  & Existence             & OCR\_Know\_Spat      & Image-Text Understanding  & Instance Identity \\
6  & Scene                 & Rec                  & Fine-grained Perception (Cross-instance)   & Scene Understanding \\
7  & OCR                   & OCR\_Gen\_Spat       & Image Style               & Instance Counting \\
8  & Poster                & Rec\_Know\_Gen       & Physical Relation         & Spatial Relation \\
9  & Color                 & Rec\_Know            & Image Scene               & Visual Reasoning \\
10 & Celebrity             & Rec\_OCR\_Gen\_Spat  & Celebrity Recognition     & — \\
11 & Landmark              & OCR                  & Identity Reasoning        & — \\
12 & Artwork               & Rec\_OCR\_Gen        & Image Emotion             & — \\
13 & Commonsense Reasoning & Rec\_OCR             & Image Topic               & — \\
14 & —                     & —                    & Natural Relation          & — \\
15 & —                     & —                    & Object Localization       & — \\
16 & —                     & —                    & OCR                       & — \\
17 & —                     & —                    & Social Relation           & — \\
18 & —                     & —                    & Attribute Recognition     & — \\
19 & —                     & —                    & Coarse Perception         & — \\
20 & —                     & —                    & Fine-grained Perception (Single-instance)   & — \\
21 & —                     & —                    & Relation Reasoning        & — \\
22 & —                     & —                    & Action Recognition        & — \\
23 & —                     & —                    & Function Reasoning        & — \\
24 & —                     & —                    & Future Prediction         & — \\
25 & —                     & —                    & Logical Reasoning         & — \\
\bottomrule
\end{tabular}
\caption{\small Subtask-to-Index Mapping Table. Each column corresponds to the substask taxonomy defined in MME \cite{fu2024mmecomprehensiveevaluationbenchmark}, MMVet~\cite{yu2023mmvetevaluatinglargemultimodal}, MMBench~\cite{liu2024mmbenchmultimodalmodelallaround} and SEEDBench~\cite{li2023seedbenchbenchmarkingmultimodalllms}.}
\label{appendix:subtasks2index}
\end{table*}
\end{document}